\documentclass{article}

\PassOptionsToPackage{numbers,compress}{natbib}

\usepackage[main,final]{neurips_2026}

\usepackage[utf8]{inputenc}
\usepackage[T1]{fontenc}
\usepackage[hidelinks]{hyperref}
\usepackage{url}
\usepackage{booktabs}
\usepackage{amsmath,amsfonts,bm,amssymb}
\usepackage{mathtools}
\usepackage{nicefrac}
\usepackage{microtype}
\usepackage{xcolor}
\usepackage{graphicx}
\usepackage{float}
\usepackage{afterpage}
\usepackage{subcaption}
\usepackage{array}
\usepackage{multirow}
\usepackage{makecell}
\usepackage{tabularx}
\usepackage{longtable}
\usepackage{siunitx}
\usepackage{adjustbox}
\usepackage{enumitem}
\usepackage[normalem]{ulem}
\usepackage{colortbl}
\usepackage{tcolorbox}
\graphicspath{{image/}}

\renewcommand{\answerYes}[1][]{\textcolor{black}{[Yes]#1}}
\renewcommand{\answerNo}[1][]{\textcolor{black}{[No]#1}}
\renewcommand{\answerNA}[1][]{\textcolor{black}{[N/A]#1}}
\renewcommand{\answerTODO}[1][]{\textcolor{black}{\bfseries [TODO]}}
\renewcommand{\justificationTODO}[1][]{\textcolor{black}{\bfseries [TODO]}}

\title{LEAF: Language-EEG Aligned Foundation Model for Brain-Computer Interfaces}

\author{%
\small\bfseries Muyun Jiang$^{1,*}$ \hspace{0.3em}%
Shuailei Zhang$^{1,*}$ \hspace{0.3em}%
Zhenjie Yang$^{2,*}$ \hspace{0.3em}%
Wu Mengjun$^1$ \hspace{0.3em}%
Wei Zhang$^1$ \hspace{0.3em}%
Weibang Jiang$^3$ \\
\small\bfseries Chenyu Liu$^1$ \hspace{0.3em}%
Zhiwei Guo$^1$ \hspace{0.3em}%
Rui Liu$^1$ \hspace{0.3em}%
Shangen Zhang$^4$ \hspace{0.3em}%
Yong Li$^{1,5}$ \hspace{0.3em}%
Yi Ding$^{1,\dagger}$ \hspace{0.3em}%
Cuntai Guan$^{1,\dagger}$ \\
\small $^1$Nanyang Technological University, Singapore \\
\small $^2$The University of Hong Kong, Hong Kong \quad
$^3$Shanghai Jiao Tong University, China \\
\small $^4$University of Science and Technology Beijing, China \quad
$^5$Southeast University, China \\
$^*$Equal contribution \quad $^\dagger$Corresponding authors
}

\begin{document}

\maketitle

\edef\restorefiguretextsep{\global\textfloatsep=\the\textfloatsep\relax}
\afterpage{%
\global\textfloatsep=12pt\relax
\afterpage{\restorefiguretextsep}
\begin{figure}[t]
    \centering
    \includegraphics[width=\linewidth]{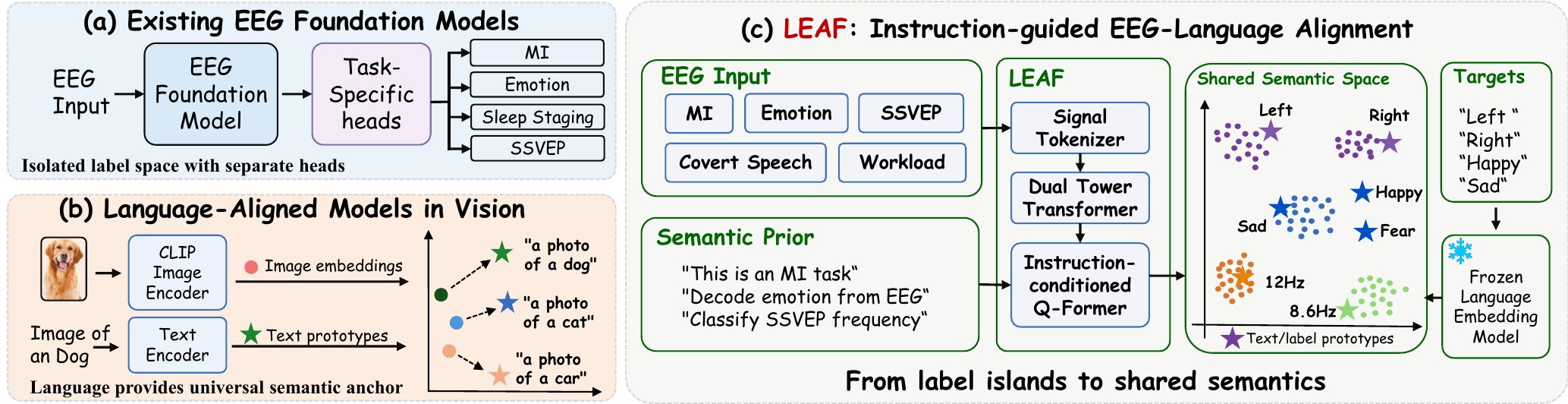}
    \caption{Motivation for LEAF. (a) EEG foundation models use task-specific heads and discrete labels. (b) CLIP aligns images with textual prototypes. (c) LEAF uses task instructions to align EEG representations with textual label prototypes.}
    \label{fig:leaf_motivation}
\end{figure}%
}

\begin{abstract}
Electroencephalography (EEG) foundation models learn transferable representations for brain--computer interfaces, but existing approaches treat tasks and labels as discrete targets and fail to use the semantic structure of natural-language task instructions to guide representation learning. We present \textbf{LEAF}, a foundation model for \textbf{EEG--Language Alignment with Semantic Task Instruction and Querying}. LEAF integrates task-aware semantic guidance to produce structured and linguistically aligned EEG embeddings, thereby enhancing decoding robustness and transferability.
In the EEG pretraining stage, we introduce a joint \textbf{Spectral--Temporal Reconstruction (STR)} framework that captures the coupled spectral rhythms and temporal dynamics of EEG signals. STR applies randomized spectral perturbation to enhance frequency robustness and uses two complementary temporal objectives to learn both contextual and sequential structure. In the EEG-Language alignment stage, we propose the \textbf{Instruction-conditioned Q-Former (IQF)}. This query-based cross-attention transformer injects instruction embeddings into EEG tokens and achieves semantic alignment with textual label embeddings through learnable queries.
\textcolor{black}{We evaluate LEAF on 16 downstream datasets spanning motor imagery, emotion recognition, steady-state visual evoked potentials, covert speech, and mental-workload decoding.} \textcolor{black}{\textcolor{black}{Under supervised finetuning,} LEAF achieves the best balanced accuracy on 12 of the 16 datasets and the strongest average performance across the evaluated paradigms.} \textcolor{black}{Further analyses show that instruction conditioning reduces intra-class variation, increases inter-class separation, and aligns EEG representations more closely with textual class prototypes. These findings support the use of task instructions as semantic priors for transferable EEG decoding. Code is available at \url{https://github.com/Jiang-Muyun/LEAF}.}
\end{abstract}

\section{Introduction}

Electroencephalography (EEG) measures brain activity noninvasively with millisecond-level temporal resolution, supporting brain--computer interface (BCI) applications such as motor imagery (MI) decoding, emotion recognition, and steady-state visual evoked potential (SSVEP) classification. Learning representations that transfer across these applications is challenging because EEG has a low signal-to-noise ratio, is nonstationary, and varies substantially across subjects, datasets, and tasks \cite{edelman2024non}. These challenges have motivated EEG foundation models (EEG-FMs), which use large-scale pretraining to learn reusable representations. Their broader promise is to share knowledge across BCI tasks, making the way these models represent both signals and tasks central to their design.

Realizing this promise requires looking beyond the EEG encoder to how downstream tasks are specified. Models such as EEGPT \cite{wang2024eegpt}, LaBraM \cite{jiang2024large}, CBraMod \cite{wang2024cbramod}, and REVE \cite{ouahidi2025reve} pretrain transformer encoders with masked or contrastive objectives and use task-specific classification heads supervised by discrete labels. Although this approach transfers signal representations, labels such as 0 and 1 do not convey the meaning of the corresponding classes or their relationship to the task. This leaves a source of transferable knowledge unused and motivates our first question: \textbf{Q1: How can EEG foundation models exploit the semantics of tasks and labels?} Natural-language instructions offer a way to describe the decoding task, while textual class descriptions provide semantic information that discrete label indices alone cannot express.

Using this semantic information requires a mechanism that connects language to EEG features. NeuroLM \cite{jiang2024neurolm} takes an important step in this direction by converting EEG signals into discrete neural tokens, aligning the tokenizer space with text through adversarial training, and performing multi-task instruction tuning within an autoregressive language model. Its EEG--language coupling is thus mediated through discrete tokenization and sequence modeling, leaving a complementary question: \textbf{Q2: Can task instructions directly guide continuous EEG representations toward semantically meaningful class structure?} We investigate this question by using instruction embeddings to condition EEG features and textual label embeddings as class prototypes in a shared space. This formulation allows task semantics to guide feature extraction and label semantics to guide alignment, linking the description of a task to the representation used to solve it.

For this instruction-guided alignment to be effective, the EEG features must also retain the signal structure needed to distinguish the described classes. EEG information is expressed through coupled spectral rhythms and temporal dynamics at multiple timescales, including event-related responses and slower contextual fluctuations \cite{wairagkar2021dynamics, li2022eeg}. Language conditioning alone does not provide an explicit objective for learning these properties. This leads to our third question: \textbf{Q3: How can spectral--temporal EEG pretraining support instruction-guided semantic alignment?} Addressing this question connects the two requirements of the model: learning informative signal representations and adapting those representations to the semantics of each decoding task.

We address these questions with \textbf{LEAF}, a foundation model that combines spectral--temporal EEG pretraining with instruction-conditioned EEG--language alignment. Its \textbf{Spectral--Temporal Reconstruction (STR)} module integrates randomized frequency suppression with bidirectional contextual reconstruction and causally masked reconstruction to learn the EEG features used in alignment. Its \textbf{Instruction-conditioned Q-Former (IQF)} then modulates these features with task instructions, refines them through multi-layer query-based cross-attention, and aligns the resulting representations with textual label prototypes through cross-entropy. Together, these components connect signal-level representation learning to task- and label-level semantic guidance. Fig.~\ref{fig:leaf_motivation} summarizes the motivation and the EEG--language alignment pipeline.

We make three contributions.
\begin{itemize}[noitemsep,topsep=0pt,leftmargin=*,itemsep=2pt]
\item We propose an instruction-conditioned EEG--language alignment framework that couples joint spectral--temporal reconstruction with query-based cross-modal refinement, replacing coarse distribution matching with fine-grained semantic guidance.
\item \textcolor{black}{We show that LEAF's EEG encoder, adapted with dataset-specific classification heads, attains state-of-the-art average performance across 16 downstream EEG datasets spanning motor imagery, emotion, SSVEP, covert speech, and mental-workload decoding.}
\item \textcolor{black}{We show that explicit task instructions act as semantic priors that organize EEG representations into more compact, class-discriminative groups within the evaluated tasks.}
\end{itemize}
\section{Method}
\label{sec:method}

\afterpage{%
\begin{figure*}[t]
\centering
\includegraphics[width=\textwidth]{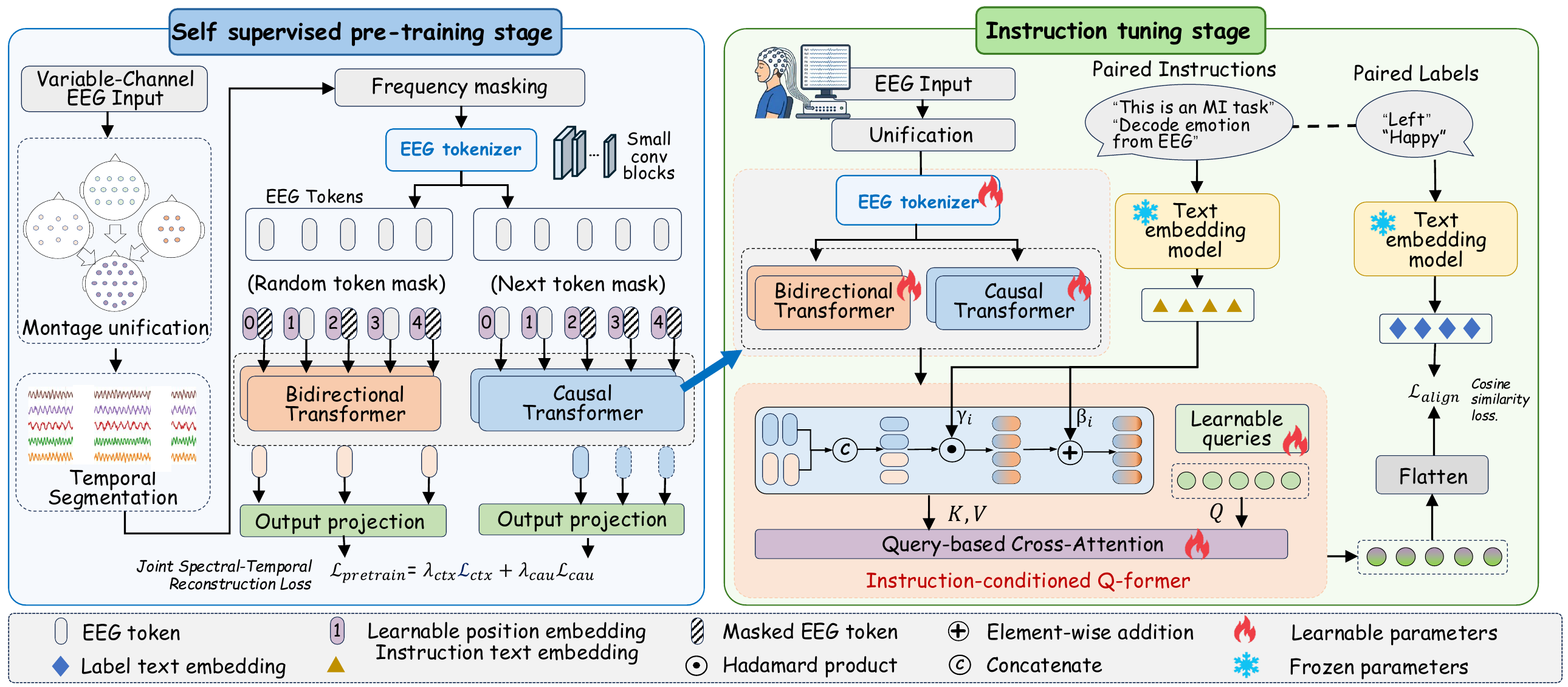}
\caption{LEAF architecture. Left: Spectral--temporal pretraining combines frequency masking with bidirectional and causal reconstruction. Right: The instruction-conditioned Q-Former aligns EEG features with textual label prototypes from a frozen text encoder.}
\label{fig:LEAF}
\end{figure*}%
}
In this section, we introduce the design of \textbf{LEAF}, our proposed EEG--Language foundation model. LEAF is trained in two stages: an EEG pretraining stage, where a joint spectral--temporal objective encourages frequency-aware and temporally predictive EEG representations, and a multi-task instruction tuning stage, where EEG embeddings are conditioned on task instructions and aligned with textual targets to improve decoding performance across diverse tasks. An overview of the LEAF architecture is shown in Fig.~\ref{fig:LEAF}.
\subsection{Pretraining with Joint Spectral--Temporal Reconstruction}
EEG signals combine spectral rhythms with temporal dynamics that exhibit both causal transitions and broader contextual fluctuations. To learn representations that capture these spectral--temporal relationships during pretraining, we propose a joint spectral-temporal reconstruction module (STR). STR is a two-stream pretraining architecture built upon a shared spectral backbone. The backbone introduces randomized frequency suppression to enforce spectral invariance, while two complementary temporal branches operate in parallel to capture contextual and sequential dependencies.
{\color{black}
Let $X\in \mathbb{R}^{C\times T}$ denote one EEG trial with $C$ channels and $T$ time points. We divide the trial into $N=\lfloor T/t\rfloor$ non-overlapping windows of length $t$,
\begin{equation}
x_i=X_{:,\,(i-1)t+1:it}\in\mathbb{R}^{C\times t},
\qquad i=1,\ldots,N,
\end{equation}
and denote the retained signal by $X^{(N)}=[x_1,\ldots,x_N]\in\mathbb{R}^{C\times Nt}$. Here, $i$ always indexes windows, whereas $N$ denotes the number of windows in the trial.
}
All EEG inputs are spatially interpolated to a standardized 65-channel 10--10 montage before tokenization; the electrode layout is provided in Appendix~\ref{appendix:parameter}.

\paragraph{Spectral masking backbone}
The spectral masking backbone encourages the encoder to learn frequency-aware features across paradigms. We first suppress a randomly chosen frequency band in each segment before tokenization.
For $x_i$, compute its spectrum $X_{f,i}=\mathrm{FFT}(x_i)$ and randomly select a band $[f_{\min}, f_{\max}]$ of width $f_{\text{band}}$ to remove, producing a masked spectrum
$\widetilde{X}_{f,i} = \mathcal{M}_{[f_{\min},f_{\max}]}(X_{f,i})$. Then we conduct an inverse transform to get the perturbed signal via
$\widetilde{x}_i = \mathrm{iFFT}(\widetilde{X}_{f,i})$.
This encourages invariance to the loss of localized spectral components and complements the dual spectral--temporal objectives.
{\color{black}
We adopt a lightweight tokenizer consisting of a temporal convolution, a spatial convolution, batch normalization, and pooling. It maps each perturbed window $\widetilde{x}_i$ to one vector $\mathbf{u}_i$, rather than to another token sequence:
\begin{equation}
\begin{aligned}
\mathbf{u}_i
&=\mathrm{vec}\!\left(\mathrm{Pool}\big(\mathrm{BatchNorm}(\mathrm{Conv}_S(\mathrm{Conv}_T(\widetilde{x}_i)))\big)\right)
\in\mathbb{R}^{d_u},\\
\mathbf{U}&=[\mathbf{u}_1,\ldots,\mathbf{u}_N]^{\top}\in\mathbb{R}^{N\times d_u}.
\end{aligned}
\end{equation}
The two temporal branches use separate linear projections $P_{\mathrm{ctx}},P_{\mathrm{cau}}:\mathbb{R}^{d_u}\!\rightarrow\mathbb{R}^{d}$:
\begin{equation}
\mathbf{Z}^{r}=P_r(\mathbf{U})
=[\mathbf{z}^{r}_1,\ldots,\mathbf{z}^{r}_N]^{\top}
\in\mathbb{R}^{N\times d},
\qquad r\in\{\mathrm{ctx},\mathrm{cau}\}.
\end{equation}
Thus, $\mathbf{u}_i$ is the tokenizer output for window $i$, $\mathbf{z}^{r}_i$ is its $d$-dimensional input to branch $r$, and the bold uppercase matrices collect all $N$ windows.
}

\paragraph{\textcolor{black}{Dual temporal masking branches}}
{\color{black}
EEG temporal dynamics contain both bidirectional contextual dependencies and temporally ordered dependencies. STR therefore uses two independent transformer branches that share the tokenizer but have separate input projections, transformer parameters, and signal decoders.

For the contextual branch, we sample a set of window indices $\mathcal{M}\subseteq\{1,\ldots,N\}$ and replace $\mathbf{z}^{\mathrm{ctx}}_i$ by a learned mask vector for every $i\in\mathcal{M}$. With positional embeddings $\mathbf{P}_{1:N}$, the bidirectional transformer produces
\begin{equation}
\mathbf{H}^{\mathrm{ctx}}
=\mathcal{T}_{\mathrm{ctx}}\!\left(\mathrm{Mask}(\mathbf{Z}^{\mathrm{ctx}},\mathcal{M})+\mathbf{P}_{1:N}\right)
\in\mathbb{R}^{N\times d}.
\end{equation}
Its decoder $g_{\mathrm{ctx}}:\mathbb{R}^{d}\rightarrow\mathbb{R}^{C\times t}$ maps each contextual state back to one EEG window. Concatenating the decoded windows gives $\widehat{X}^{\mathrm{ctx}}=[g_{\mathrm{ctx}}(\mathbf{h}^{\mathrm{ctx}}_1),\ldots,g_{\mathrm{ctx}}(\mathbf{h}^{\mathrm{ctx}}_N)]\in\mathbb{R}^{C\times Nt}$, and the implementation applies mean-squared reconstruction error over the complete retained signal:
\begin{equation}
\mathcal{L}_{\mathrm{ctx}}
=\frac{1}{CNt}\left\|\widehat{X}^{\mathrm{ctx}}-X^{(N)}\right\|_F^2.
\end{equation}

For the causal branch, a learned beginning-of-sequence vector $\mathbf{b}$ is prepended and a triangular attention mask prevents every position from attending to future windows:
\begin{equation}
\mathbf{H}^{\mathrm{cau}}
=\mathcal{T}_{\mathrm{cau}}\!\left([\mathbf{b};\mathbf{Z}^{\mathrm{cau}}]+\mathbf{P}_{1:N+1};\,\mathbf{A}_{\mathrm{cau}}\right)
\in\mathbb{R}^{(N+1)\times d},
\end{equation}
where $(\mathbf{A}_{\mathrm{cau}})_{ij}=-\infty$ for $j>i$ and $0$ otherwise. The state $\mathbf{h}^{\mathrm{cau}}_{i+1}$ can therefore use only the current and preceding window tokens $\{\mathbf{z}^{\mathrm{cau}}_1,\ldots,\mathbf{z}^{\mathrm{cau}}_i\}$. A separate decoder $g_{\mathrm{cau}}$ reconstructs each original window, yielding $\widehat{X}^{\mathrm{cau}}=[g_{\mathrm{cau}}(\mathbf{h}^{\mathrm{cau}}_2),\ldots,g_{\mathrm{cau}}(\mathbf{h}^{\mathrm{cau}}_{N+1})]$, with
\begin{equation}
\mathcal{L}_{\mathrm{cau}}
=\frac{1}{CNt}\left\|\widehat{X}^{\mathrm{cau}}-X^{(N)}\right\|_F^2.
\end{equation}
}

\paragraph{\textcolor{black}{Joint spectral--temporal objective}}
{\color{black}
Frequency suppression acts as input corruption shared by both branches; it does not introduce a third loss term. The complete pretraining objective is
\begin{equation}
\mathcal{L}_{pretrain}=\lambda_{ctx}\mathcal{L}_{ctx}+\lambda_{cau}\mathcal{L}_{cau},
\end{equation}
where $\lambda_{\mathrm{ctx}}$ and $\lambda_{\mathrm{cau}}$ are set to $1$ by default. During instruction tuning and inference, random frequency suppression and token masking are disabled, while the causal attention mask is retained. The contextual and causal branches each produce $N$ encoded states from the uncorrupted token sequence, and their outputs are concatenated along the sequence dimension to form $\mathbf{m}\in\mathbb{R}^{2N\times d}$, which is passed to the IQF.
}

\subsection{EEG-Language Alignment with Multi-task Instruction Tuning}
Multi-task instruction tuning bridges EEG and language by conditioning EEG embeddings on task instructions and aligning them with textual label prototypes, establishing fine-grained semantic correspondence beyond coarse distribution-level matching.

\paragraph{Instruction-conditioned EEG-Language Interaction}
Previous EEG--language models such as NeuroLM achieve alignment through adversarial domain matching between neural and textual representations. In contrast, we proposed the Instruction-conditioned Q-Former (IQF). It achieves EEG--language alignment in a constructive and interpretable manner by conditioning neural representations on semantics and guiding cross-modal interaction.
\textcolor{black}{Using the notation above, $\mathbf{m}\in\mathbb{R}^{2N\times d}$ is the concatenated output of the contextual and causal EEG branches. \textcolor{black}{Given instruction text $s_{\mathrm{ins}}$, we obtain its $\ell_2$-normalized embedding via a frozen language embedding model:}}
\begin{equation}
{\color{black}
\mathbf{e}_{\mathrm{ins}} = \frac{f_{\text{text}}(s_{\mathrm{ins}})}{\lVert f_{\text{text}}(s_{\mathrm{ins}}) \rVert_2},
}
\label{eq:text-enc}
\end{equation}
\textcolor{black}{For dataset $D$ with candidate labels $\mathcal{C}_D$, task-and-target conditioning combines this instruction embedding with the normalized mean of its fixed label prototypes $\overline{\mathbf{e}}_c$ (defined below). Writing $\operatorname{norm}(\mathbf{v})=\mathbf{v}/\|\mathbf{v}\|_2$, we use}
\begin{equation}
{\color{black}
\mathbf{a}_D=\operatorname{norm}\!\left(\frac{1}{|\mathcal{C}_D|}\sum_{c\in\mathcal{C}_D}\overline{\mathbf{e}}_c\right),
\qquad
\mathbf{e}_{\mathrm{cond}}=\operatorname{norm}\!\left(\frac{\mathbf{e}_{\mathrm{ins}}+\mathbf{a}_D}{2}\right).
}
\label{eq:instruction-blend}
\end{equation}
\textcolor{black}{The mean uses all candidate labels, not the trial's ground-truth label. During instruction tuning, we sample task paraphrases, substituting a generic default instruction with probability $0.1$. Independently, we apply the candidate-label blend with probability $0.9$; otherwise, $\mathbf{e}_{\mathrm{cond}}=\mathbf{e}_{\mathrm{ins}}$.}

To fuse this conditioning prior with the EEG embedding space, we employ a Feature-wise Linear Modulation (FiLM) operator~\cite{perez2017learning}, which parameterizes an affine transformation. \textcolor{black}{Specifically, the modulation parameters $(\boldsymbol{\gamma}, \boldsymbol{\beta})$ are derived from the conditioning embedding $\mathbf{e}_{\mathrm{cond}}$ via a nonlinear projection:}
\begin{equation}
{\color{black}(\boldsymbol{\gamma}, \boldsymbol{\beta}) \;=\; \tanh\!\big( \mathbf{W}_{\gamma\beta}\,\mathbf{e}_{\mathrm{cond}} + \mathbf{b}_{\gamma\beta} \big),}
\label{eq:film-proj}
\end{equation}
where $\mathbf{W}_{\gamma\beta}\in\mathbb{R}^{2d\times k}$ and $\mathbf{b}_{\gamma\beta}\in\mathbb{R}^{2d}$. We split $(\boldsymbol{\gamma}, \boldsymbol{\beta})$ into two $d$-dimensional vectors, $\boldsymbol{\gamma}\in\mathbb{R}^{d}$ and $\boldsymbol{\beta}\in\mathbb{R}^{d}$. For element-wise modulation, these vectors are broadcast from $(d)$ to $(2N\times d)$ so that each latent dimension of every token is modulated by the same semantic prior.
The instruction-conditioned representation is then obtained by
\begin{equation}
\tilde{\mathbf{m}} = (1 + \boldsymbol{\gamma}) \odot \mathbf{m} + \boldsymbol{\beta},
\label{eq:film-mod}
\end{equation}

where $\odot$ denotes element-wise multiplication. This formulation ensures that $\mathbf{m}$ is shaped by instruction semantics rather than generic alignment. The \textcolor{black}{conditioning embedding $\mathbf{e}_{\mathrm{cond}}$} biases the EEG latent space toward task-relevant features, producing representations that are both aligned with textual targets $\mathbf{e}_{\text{tgt}}$ and regularized on an instruction-informed manifold for improved semantic fidelity and generalization.

{\color{black}
To extract instruction-relevant neural features from the modulated EEG sequence, IQF uses $N_q$ shared learnable queries $\mathbf{Q}_0\in\mathbb{R}^{N_q\times d}$. We first add positional embeddings to the instruction-modulated EEG tokens,
\begin{equation}
\mathbf{M}=\widetilde{\mathbf{m}}+\mathbf{P}_{1:2N}\in\mathbb{R}^{2N\times d},
\end{equation}
and pass the queries through a stack of $L$ Q-Former layers. Each layer contains query self-attention, cross-attention from the queries to the fixed EEG token sequence $\mathbf{M}$, and a feed-forward network, all implemented with pre-normalization and residual connections. For $\ell=1,\ldots,L$,
\begin{equation}
\begin{aligned}
\mathbf{S}_{\ell}
&=\mathbf{Q}_{\ell-1}
+\mathrm{MHA}^{\ell}_{\mathrm{self}}\!\left(
\mathrm{LN}^{\ell}_{1}(\mathbf{Q}_{\ell-1}),
\mathrm{LN}^{\ell}_{1}(\mathbf{Q}_{\ell-1}),
\mathrm{LN}^{\ell}_{1}(\mathbf{Q}_{\ell-1})\right),\\
\mathbf{C}_{\ell}
&=\mathbf{S}_{\ell}
+\mathrm{MHA}^{\ell}_{\mathrm{cross}}\!\left(
\mathrm{LN}^{\ell}_{2}(\mathbf{S}_{\ell}),\mathbf{M},\mathbf{M}\right),\\
\mathbf{Q}_{\ell}
&=\mathbf{C}_{\ell}
+\mathrm{FFN}^{\ell}\!\left(\mathrm{LN}^{\ell}_{3}(\mathbf{C}_{\ell})\right).
\end{aligned}
\end{equation}
The three arguments of each multi-head attention operator are its query, key, and value, respectively. Thus, self-attention first allows the learnable queries to exchange information, and cross-attention then lets every query retrieve instruction-conditioned EEG evidence at every layer, rather than only once. After the final layer, the queries are normalized, flattened, and projected into the $k$-dimensional text-embedding space:
\begin{equation}
\mathbf{h}=f_{\mathrm{out}}\!\left(\mathrm{vec}(\mathrm{LN}_{f}(\mathbf{Q}_{L}))\right)\in\mathbb{R}^{k},
\qquad
\hat{\mathbf{h}}=\frac{\mathbf{h}}{\|\mathbf{h}\|_2}.
\end{equation}
}

\paragraph{Textual Prototype-based Semantic Alignment}
{\color{black}
\textcolor{black}{Let $\mathcal{V}=\bigcup_D\mathcal{C}_D$ denote the global vocabulary of distinct textual labels across all instruction-tuning datasets. For every class $c\in\mathcal{V}$, we encode its textual label $s_{\mathrm{tgt}}^c$ with the same frozen language encoder used for instructions and normalize the result to obtain a fixed prototype:}
\begin{equation}
\overline{\mathbf{e}}_{c}
=\frac{f_{\mathrm{text}}(s_{\mathrm{tgt}}^c)}
{\left\|f_{\mathrm{text}}(s_{\mathrm{tgt}}^c)\right\|_2}
\in\mathbb{R}^{k}.
\end{equation}
Using the same encoder for both instructions and labels ensures that they are represented in a shared semantic space.

Because both $\hat{\mathbf{h}}_b$ and $\overline{\mathbf{e}}_c$ are $\ell_2$-normalized, their dot product is their cosine similarity. For sample $b$ and every candidate class $c$, we first compute the prototype score
\begin{equation}
\ell_{b,c}
=\hat{\mathbf{h}}_b^{\top}\overline{\mathbf{e}}_c
=\cos\!\left(\hat{\mathbf{h}}_b,\overline{\mathbf{e}}_c\right).
\end{equation}

\textcolor{black}{During instruction tuning, cross-entropy uses all prototypes in the global vocabulary $\mathcal{V}$, including labels from other datasets; $y_b$ indexes the ground-truth label in this vocabulary:}
\begin{equation}
{\color{black}
\mathcal{L}_{\mathrm{align}}
=-\frac{1}{B}\sum_{b=1}^{B}
\log
\frac{\exp(\ell_{b,y_b})}
{\sum_{c\in\mathcal{V}}\exp(\ell_{b,c})}.
}
\end{equation}
Consequently, minimizing $\mathcal{L}_{\mathrm{align}}$ increases similarity to the ground-truth prototype while decreasing relative similarity to every incorrect prototype through the softmax denominator. The loss therefore provides both positive alignment and explicit negative-class competition in the shared EEG--language space.
}

\subsection{Direct Inference}
After multi-task instruction tuning, LEAF is evaluated on each downstream dataset's held-out test split with all model weights fixed. \textcolor{black}{Each EEG trial is processed by the EEG encoder and IQF using $\mathbf{e}_{\mathrm{cond}}$, yielding the instruction-conditioned embedding $\hat{\mathbf{h}}$. Task-and-target inference uses the blend in Eq.~\eqref{eq:instruction-blend}; task-only and no-instruction inference use the task instruction or generic default embedding, respectively, without candidate-label blending. No stochastic instruction substitution or blend dropout is applied at inference.} The frozen text encoder provides a prototype $\mathbf{e}_{\text{tgt}}^{\,y}$ for each candidate class label. The predicted class is the prototype with the highest cosine similarity:
\[
{\color{black}\hat{y} = \arg\max_{y\in\mathcal{C}_D} \cos(\hat{\mathbf{h}}, \mathbf{e}_{\text{tgt}}^{\,y}),}
\]
\textcolor{black}{where $\mathcal{C}_D$ is the candidate class set of the evaluated dataset $D$, rather than the global training vocabulary $\mathcal{V}$.} This procedure requires no additional classifier or dataset-specific parameter updates. Because instruction tuning uses the training splits of these downstream datasets, this setting evaluates direct decoding on held-out examples of previously instruction-tuned tasks.

\subsection{\textcolor{black}{Task-Specific Finetuning with an MLP Head}}
For supervised adaptation, we initialize the EEG encoder from the same instruction-tuned checkpoint and attach a randomly initialized, dataset-specific multilayer perceptron (MLP). The MLP takes the flattened, concatenated token outputs of the contextual and causal encoder branches. All EEG encoder parameters used in this prediction path and the MLP are jointly optimized on the target dataset's training split using cross-entropy loss. The IQF and textual class prototypes are excluded from this classification path. At test time, the adapted encoder and MLP process each held-out EEG trial, and the class with the largest classifier logit is selected. Thus, \textcolor{black}{MLP-head finetuning} measures supervised transfer of the instruction-tuned EEG encoder, whereas direct inference measures decoding through the fixed EEG--language alignment. Appendix~\ref{appendix:qformer_finetuning} illustrates and compares this strategy with Q-Former finetuning using cosine-similarity logits and cross-entropy, which retains the existing model without adding trainable parameters.

\section{Experiments}

\subsection{Dataset}

\textcolor{black}{For pretraining, we use six datasets: Stieger2021~\cite{stieger2021mindfulness}, SEED-FRA and SEED-GER~\cite{liu2022identifying}, SEED-SD~\cite{li2025investigating}, LargeSpanish~\cite{valle2024identification}, and ThinkOutLoud~\cite{nieto2022thinking}. The processed corpus contains 325,046 segments with 65 channels at 200\,Hz, totaling approximately 885.71 hours of stored signal, including padding introduced during preprocessing. This total includes both the training and validation portions. Dataset-specific preprocessing and validation splits are detailed in Appendix~\ref{appendix:pretrain_datasets}.}

\textcolor{black}{We evaluate LEAF on 16 datasets spanning motor imagery, emotion recognition, SSVEP, covert speech, and mental-workload decoding. Table~\ref{tab:dataset_overview} summarizes their evaluation splits and textual class labels; detailed dataset descriptions are provided in Appendix~\ref{appendix:downstream_datasets}.}

\begin{table*}[!t]
\centering
\caption{Downstream datasets, evaluation splits, and textual class labels. CS: cross-subject; CT: cross-trial. The class count is shown in parentheses.}
\label{tab:dataset_overview}
\scriptsize
\setlength{\tabcolsep}{3pt}
\renewcommand{\arraystretch}{1.05}
\begin{tabularx}{\textwidth}{@{}r l c l l l >{\raggedright\arraybackslash}X@{}}
\toprule
\textbf{ID} & \textbf{Dataset} & \textbf{Split} & \textbf{Train} & \textbf{Validation} & \textbf{Test} & \textbf{(\#cls) Textual labels} \\
\midrule
1 & BCIC-IV-2a~\cite{tangermann2012review} & CS & Subj.\ 1--7 & 20\% of train & Subj.\ 8--9 & (4) Left, Right, Foot, Tongue \\
2 & OpenBMI-MI~\cite{lee2019eeg} & CS & Subj.\ 1--42 & 20\% of train & Subj.\ 43--54 & (2) Right, Left \\
3 & BCIC-Upperlimb~\cite{jeong20222020} & CS & Subj.\ 1--11 & 20\% of train & Subj.\ 12--15 & (3) Cylin, Sphe, Lumbrical \\
4 & SHU-MI~\cite{ma2022large} & CS & Subj.\ 1--20 & 20\% of train & Subj.\ 21--25 & (2) Right, Left \\
5 & HighGamma~\cite{schirrmeister2017deep} & CS & Subj.\ 1--10 & 20\% of train & Subj.\ 11--14 & \textcolor{black}{(2) Left, Right} \\
6 & Cho2017~\cite{cho2017eeg} & CS & Subj.\ 1--40 & 20\% of train & Subj.\ 41--49 & (2) Left, Right \\
7 & Shin2017A~\cite{shin2016open} & CS & Subj.\ 1--22 & 20\% of train & Subj.\ 23--28 & (2) Left, Right \\
8 & PhysioNet-MI~\cite{schalk2004bci2000} & CS & Subj.\ 1--80 & 20\% of train & Subj.\ 81--109 & (2) Left, Right \\
\midrule
9 & FACED~\cite{chen2023large} & CS & Subj.\ 1--100 & 20\% of train & Subj.\ 101--122 & (9) Anger, Fear, Disgust, Sad, Amusement, Inspiration, Joy, Tenderness, Neutral \\
10 & SEED~\cite{duan2013differential} & CT & Trial 1--9 & Trial 10--12 & Trial 13--15 & (3) Positive, Neutral, Negative \\
11 & SEED-IV~\cite{zheng2018emotionmeter} & CT & Trial 1--16 & Trial 17--20 & Trial 21--24 & (4) Neutral, Sad, Fear, Happy \\
12 & SEED-V~\cite{liu2021comparing} & CT & Trial 1--5 & Trial 6--10 & Trial 11--15 & (5) Disgust, Fear, Sad, Neutral, Happy \\
13 & SEED-VII~\cite{jiang2024seed} & CT & Trial 1--10 & Trial 11--15 & Trial 16--20 & (7) Happy, Surprise, Neutral, Sad, Disgust, Fear, Anger \\
\midrule
14 & OpenBMI-SSVEP~\cite{lee2019eeg} & CS & Subj.\ 1--42 & 20\% of train & Subj.\ 43--54 & (4) 12.0, 8.6, 6.6, 5.4\,Hz \\
15 & Mental Workload~\cite{zyma2019electroencephalograms} & CS & Subj.\ 0--31 & 20\% of train & Subj.\ 32--35 & (2) Resting, Workload \\
16 & BCIC-Speech~\cite{jeong20222020} & CT & Trial 1--250 & Trial 251--300 & Trial 301--350 & (5) hello, help-me, stop, thank-you, yes \\
\bottomrule
\end{tabularx}
\end{table*}

\subsection{Experimental Setup}

\begingroup
\raggedbottom
In this paper, we selected both the state-of-the-art traditional models and the EEG-FMs as baselines. For the traditional models, we selected EEGNet \cite{lawhern2018eegnet}, TSception \cite{ding2022tsception}, ST-Transformer \cite{song2021transformer} and Conformer \cite{song2022eeg}. For the EEG foundation model, we selected BIOT \cite{yang2023biot}, EEGPT \cite{wang2024eegpt}, LaBraM \cite{jiang2024large}, and CBraMod \cite{wang2024cbramod}. Implementation about the baselines can be found in Appendix \ref{appendix:baselines}. \textcolor{black}{To provide a reliable evaluation across imbalanced datasets, we adopted \textbf{balanced accuracy} and \textbf{Cohen's Kappa} as performance metrics.} EEG inputs use a sampling rate of 200\,Hz, with dataset-specific filtering and segmentation. After amplitude clipping and robust normalization, source channels are mapped to the 65-channel LEAF montage using a fixed assignment policy: signals assigned to the same template electrode are averaged, and missing electrodes are spatially interpolated. Dataset-specific preprocessing is detailed in Appendices~\ref{appendix:pretrain_datasets} and~\ref{appendix:downstream_datasets}; the montage and complete channel-mapping policy are provided in Appendix~\ref{appendix:montage_config}.

Table~\ref{tab:dataset_overview} summarizes the downstream evaluation splits and textual class labels; Appendix~\ref{appendix:downstream_datasets} specifies the training, validation, and test partitions for each dataset. Pretraining and instruction tuning are performed end-to-end. LEAF contains approximately 26.42\,M parameters, excluding the frozen text encoder and dataset-specific classification heads. Architecture and training settings are provided in Appendix~\ref{appendix:parameter}.

\subsection{Experimental Results}

\textcolor{black}{Tables~\ref{tab:mean_std} and~\ref{tab:mean_std_other} compare LEAF with eight baselines under the supervised finetuning protocol described in Section~2.4.} Balanced accuracy (B.Acc) is reported as mean~$\pm$~standard deviation over three random seeds.

For motor imagery, LEAF achieves the highest mean balanced accuracy on five of the eight datasets: BCIC-IV-2a, BCIC-Upperlimb, HighGamma, Cho2017, and Shin2017A. Its equal-weight average across the eight datasets is 70.34\%, compared with 68.87\% for CBraMod, the strongest baseline on this average. LEAF trails the best baseline on OpenBMI-MI, SHU-MI, and PhysioNet-MI, showing that the benefit of encoder transfer varies across datasets.
\begingroup
\setlength{\intextsep}{5pt}
\begin{table}[H]
\centering
\captionsetup[subtable]{font=footnotesize, skip=1pt}
\caption{\textcolor{black}{Task-specific finetuning performance on eight motor imagery datasets.}}
\label{tab:mean_std}
\renewcommand{\arraystretch}{1.0}

\begin{subtable}{1.0\textwidth}
\centering
\scriptsize
\begin{adjustbox}{width=\textwidth}
\begin{tabular}{l cc cc cc cc}
\toprule
\multirow{2}{*}{Methods}
& \multicolumn{2}{c}{BCIC-IV-2a}
& \multicolumn{2}{c}{OpenBMI-MI}
& \multicolumn{2}{c}{BCIC-Upperlimb}
& \multicolumn{2}{c}{SHU-MI} \\
\cmidrule(lr){2-3}\cmidrule(lr){4-5}\cmidrule(lr){6-7}\cmidrule(lr){8-9}
& B.Acc & \textcolor{black}{Kappa} & B.Acc & \textcolor{black}{Kappa} & B.Acc & \textcolor{black}{Kappa} & B.Acc & \textcolor{black}{Kappa} \\
\midrule
EEGNet    & \underline{0.6369$\pm$0.0097} & \underline{0.5159$\pm$0.0129} & \underline{0.8170$\pm$0.0032} & \underline{0.6340$\pm$0.0064} & 0.5281$\pm$0.0046 & 0.2922$\pm$0.0069 & 0.5665$\pm$0.0007 & 0.1330$\pm$0.0014 \\
Conformer & 0.6347$\pm$0.0118 & 0.5129$\pm$0.0157 & \textbf{0.8274$\pm$0.0027} & \textbf{0.6548$\pm$0.0054} & \underline{0.5473$\pm$0.0039} & \underline{0.3210$\pm$0.0059} & 0.5780$\pm$0.0008 & 0.1560$\pm$0.0016 \\
TScep.    & 0.6252$\pm$0.0106 & 0.5003$\pm$0.0141 & 0.6665$\pm$0.0032 & 0.3330$\pm$0.0064 & 0.5199$\pm$0.0046 & 0.2799$\pm$0.0069 & 0.5564$\pm$0.0007 & 0.1128$\pm$0.0014 \\
STTran.   & 0.5816$\pm$0.0100 & 0.4421$\pm$0.0133 & 0.7514$\pm$0.0030 & 0.5028$\pm$0.0060 & 0.5284$\pm$0.0037 & 0.2926$\pm$0.0056 & 0.5714$\pm$0.0008 & 0.1428$\pm$0.0016 \\
BIOT      & 0.5139$\pm$0.0099 & 0.3519$\pm$0.0132 & 0.5613$\pm$0.0029 & 0.1226$\pm$0.0058 & 0.4595$\pm$0.0046 & 0.1893$\pm$0.0069 & 0.5587$\pm$0.0008 & 0.1174$\pm$0.0016 \\
EEGPT     & 0.5279$\pm$0.0123 & 0.3705$\pm$0.0164 & 0.7323$\pm$0.0026 & 0.4646$\pm$0.0052 & 0.5231$\pm$0.0044 & 0.2847$\pm$0.0066 & 0.5228$\pm$0.0009 & 0.0456$\pm$0.0018 \\
LaBraM    & 0.6234$\pm$0.0125 & 0.4979$\pm$0.0167 & 0.7874$\pm$0.0032 & 0.5748$\pm$0.0064 & 0.5414$\pm$0.0037 & 0.3121$\pm$0.0056 & \underline{0.6338$\pm$0.0008} & \underline{0.2676$\pm$0.0016} \\
CBraMod   & 0.6139$\pm$0.0117 & 0.4852$\pm$0.0156 & 0.7895$\pm$0.0028 & 0.5790$\pm$0.0056 & 0.5454$\pm$0.0042 & 0.3181$\pm$0.0063 & \textbf{0.6403$\pm$0.0007} & \textbf{0.2806$\pm$0.0014} \\
\midrule
\rowcolor{blue!8}
LEAF      & \textbf{0.6381$\pm$0.0120} & \textbf{0.5175$\pm$0.0160} & 0.8134$\pm$0.0031 & 0.6268$\pm$0.0062 & \textbf{0.5483$\pm$0.0038} & \textbf{0.3225$\pm$0.0057} & 0.6134$\pm$0.0008 & 0.2268$\pm$0.0016 \\
\bottomrule
\end{tabular}
\end{adjustbox}
\end{subtable}

\vspace{2pt}

\begin{subtable}{1.0\textwidth}
\centering
\scriptsize
\begin{adjustbox}{width=\textwidth}
\begin{tabular}{l cc cc cc cc}
\multirow{2}{*}{Methods}
& \multicolumn{2}{c}{HighGamma}
& \multicolumn{2}{c}{Cho2017}
& \multicolumn{2}{c}{Shin2017A}
& \multicolumn{2}{c}{PhysioNet-MI} \\
\cmidrule(lr){2-3}\cmidrule(lr){4-5}\cmidrule(lr){6-7}\cmidrule(lr){8-9}
& B.Acc & \textcolor{black}{Kappa} & B.Acc & \textcolor{black}{Kappa} & B.Acc & \textcolor{black}{Kappa} & B.Acc & \textcolor{black}{Kappa} \\
\midrule
EEGNet    & \underline{0.7856$\pm$0.0038} & \textcolor{black}{\underline{0.5712$\pm$0.0076}} & 0.7644$\pm$0.0062 & 0.5288$\pm$0.0124 & \underline{0.7054$\pm$0.0085} & \underline{0.4108$\pm$0.0170} & 0.6953$\pm$0.0029 & 0.3906$\pm$0.0058 \\
Conformer & 0.7637$\pm$0.0036 & \textcolor{black}{0.5274$\pm$0.0072} & \underline{0.7838$\pm$0.0057} & \underline{0.5676$\pm$0.0114} & 0.6464$\pm$0.0071 & 0.2928$\pm$0.0142 & 0.6951$\pm$0.0034 & 0.3902$\pm$0.0068 \\
TScep.    & 0.6851$\pm$0.0031 & \textcolor{black}{0.3702$\pm$0.0062} & 0.7339$\pm$0.0056 & 0.4678$\pm$0.0112 & 0.5992$\pm$0.0089 & 0.1984$\pm$0.0178 & 0.6649$\pm$0.0030 & 0.3298$\pm$0.0060 \\
STTran.   & 0.7019$\pm$0.0028 & \textcolor{black}{0.4038$\pm$0.0056} & 0.7612$\pm$0.0059 & 0.5224$\pm$0.0118 & 0.6168$\pm$0.0093 & 0.2336$\pm$0.0186 & 0.6706$\pm$0.0035 & 0.3412$\pm$0.0070 \\
BIOT      & 0.5824$\pm$0.0035 & \textcolor{black}{0.1648$\pm$0.0070} & 0.5413$\pm$0.0055 & 0.0826$\pm$0.0110 & 0.5422$\pm$0.0094 & 0.0844$\pm$0.0188 & 0.4874$\pm$0.0030 & -0.0252$\pm$0.0060 \\
EEGPT     & 0.6516$\pm$0.0038 & \textcolor{black}{0.3032$\pm$0.0076} & 0.7197$\pm$0.0058 & 0.4394$\pm$0.0116 & 0.5389$\pm$0.0074 & 0.0778$\pm$0.0148 & 0.6820$\pm$0.0029 & 0.3640$\pm$0.0058 \\
LaBraM    & 0.6939$\pm$0.0036 & \textcolor{black}{0.3878$\pm$0.0072} & 0.7614$\pm$0.0053 & 0.5228$\pm$0.0106 & 0.6744$\pm$0.0073 & 0.3488$\pm$0.0146 & \textbf{0.7246$\pm$0.0028} & \textbf{0.4492$\pm$0.0056} \\
CBraMod   & 0.7684$\pm$0.0034 & \textcolor{black}{0.5368$\pm$0.0068} & 0.7439$\pm$0.0059 & 0.4878$\pm$0.0118 & 0.6861$\pm$0.0072 & 0.3722$\pm$0.0144 & \underline{0.7219$\pm$0.0037} & \underline{0.4438$\pm$0.0074} \\
\midrule
\rowcolor{blue!8}
LEAF      & \textbf{0.7982$\pm$0.0035} & \textcolor{black}{\textbf{0.5964$\pm$0.0070}} & \textbf{0.7908$\pm$0.0060} & \textbf{0.5816$\pm$0.0120} & \textbf{0.7256$\pm$0.0088} & \textbf{0.4512$\pm$0.0176} & 0.6992$\pm$0.0035 & 0.3984$\pm$0.0070 \\
\bottomrule
\end{tabular}
\end{adjustbox}
\end{subtable}

\end{table}
\endgroup

\endgroup

For emotion recognition, LEAF obtains the highest mean balanced accuracy on FACED, SEED-IV, SEED-V, and SEED-VII, with gains of 0.32, 0.25, 0.97, and 0.45 percentage points over the strongest baseline on each dataset. On SEED, it reaches 70.11\%, below CBraMod's 71.02\%. The five-dataset averages are 49.88\% for LEAF and 49.67\% for CBraMod. These modest numerical differences should be interpreted alongside the reported variation across seeds.

Across the remaining paradigms, LEAF achieves 94.62\% on OpenBMI-SSVEP, 54.53\% on BCIC-Speech, and 64.93\% on Mental Workload, exceeding the strongest baseline by 0.32, 1.39, and 0.13 percentage points, respectively. Across all 16 datasets, LEAF ranks first in mean balanced accuracy on 12 datasets and achieves an equal-weight average of 64.14\%, compared with 61.97\% for CBraMod. These results support the transferability of the instruction-tuned EEG encoder under \textcolor{black}{supervised finetuning}.
\begin{table}[H]
\vspace{-7pt} 
\centering
\captionsetup[subtable]{font=footnotesize, skip=1pt}
\caption{\textcolor{black}{Task-specific finetuning performance on eight datasets spanning emotion recognition, SSVEP, covert speech, and mental workload.}}
\label{tab:mean_std_other}
\renewcommand{\arraystretch}{1.0}

\begin{subtable}{1.0\textwidth}
\centering
\scriptsize
\begin{adjustbox}{width=\textwidth}
\begin{tabular}{l cc cc cc cc}
\toprule
\multirow{2}{*}{Methods}
& \multicolumn{2}{c}{FACED}
& \multicolumn{2}{c}{SEED}
& \multicolumn{2}{c}{SEED-IV}
& \multicolumn{2}{c}{SEED-V} \\
\cmidrule(lr){2-3}\cmidrule(lr){4-5}\cmidrule(lr){6-7}\cmidrule(lr){8-9}
& B.Acc & \textcolor{black}{Kappa} & B.Acc & \textcolor{black}{Kappa} & B.Acc & \textcolor{black}{Kappa} & B.Acc & \textcolor{black}{Kappa} \\
\midrule
EEGNet    & 0.4271$\pm$0.0019 & 0.3552$\pm$0.0021 & 0.5337$\pm$0.0122 & 0.3001$\pm$0.0183 & 0.3651$\pm$0.0102 & 0.1421$\pm$0.0128 & 0.2932$\pm$0.0102 & 0.1156$\pm$0.0127 \\
Conformer & 0.4943$\pm$0.0019 & 0.4308$\pm$0.0021 & 0.6254$\pm$0.0125 & 0.4376$\pm$0.0187 & 0.4094$\pm$0.0112 & 0.1978$\pm$0.0142 & 0.3060$\pm$0.0112 & 0.1315$\pm$0.0139 \\
TScep.    & 0.2056$\pm$0.0022 & 0.1062$\pm$0.0025 & 0.6369$\pm$0.0155 & 0.4548$\pm$0.0232 & 0.4063$\pm$0.0078 & 0.1939$\pm$0.0099 & 0.3637$\pm$0.0078 & 0.2032$\pm$0.0097 \\
STTran.   & 0.3791$\pm$0.0023 & 0.3012$\pm$0.0026 & 0.5882$\pm$0.0079 & 0.3818$\pm$0.0118 & 0.3616$\pm$0.0072 & 0.1377$\pm$0.0090 & 0.2244$\pm$0.0072 & 0.0302$\pm$0.0089 \\
BIOT      & 0.1711$\pm$0.0022 & 0.0674$\pm$0.0025 & 0.6674$\pm$0.0118 & 0.5006$\pm$0.0177 & 0.4141$\pm$0.0187 & 0.2037$\pm$0.0237 & 0.3045$\pm$0.0187 & 0.1296$\pm$0.0232 \\
EEGPT     & 0.3346$\pm$0.0025 & 0.2512$\pm$0.0028 & 0.5054$\pm$0.0105 & 0.2577$\pm$0.0157 & 0.3202$\pm$0.0144 & 0.0862$\pm$0.0178 & 0.2253$\pm$0.0144 & 0.0313$\pm$0.0178 \\
LaBraM    & 0.5457$\pm$0.0021 & 0.4886$\pm$0.0024 & \underline{0.7083$\pm$0.0107} & \underline{0.5619$\pm$0.0161} & 0.4415$\pm$0.0138 & 0.2385$\pm$0.0176 & 0.4010$\pm$0.0138 & 0.2495$\pm$0.0172 \\
CBraMod   & \underline{0.5787$\pm$0.0021} & \underline{0.5257$\pm$0.0024} & \textbf{0.7102$\pm$0.0089} & \textbf{0.5648$\pm$0.0134} & \underline{0.4605$\pm$0.0097} & \underline{0.2628$\pm$0.0124} & \underline{0.4029$\pm$0.0138} & \underline{0.2519$\pm$0.0172} \\
\midrule
\rowcolor{blue!8}
LEAF      & \textbf{0.5819$\pm$0.0023} & \textbf{0.5293$\pm$0.0026} & 0.7011$\pm$0.0068 & 0.5511$\pm$0.0102 & \textbf{0.4630$\pm$0.0034} & \textbf{0.2660$\pm$0.0044} & \textbf{0.4126$\pm$0.0023} & \textbf{0.2640$\pm$0.0029} \\
\bottomrule
\end{tabular}
\end{adjustbox}
\end{subtable}

\vspace{2pt}

\begin{subtable}{1.0\textwidth}
\centering
\scriptsize
\begin{adjustbox}{width=\textwidth}
\begin{tabular}{l cc cc cc cc}
\multirow{2}{*}{Methods}
& \multicolumn{2}{c}{SEED-VII}
& \multicolumn{2}{c}{OpenBMI-SSVEP}
& \multicolumn{2}{c}{BCIC-Speech}
& \multicolumn{2}{c}{Mental Workload} \\
\cmidrule(lr){2-3}\cmidrule(lr){4-5}\cmidrule(lr){6-7}\cmidrule(lr){8-9}
& B.Acc & \textcolor{black}{Kappa} & B.Acc & \textcolor{black}{Kappa} & B.Acc & \textcolor{black}{Kappa} & B.Acc & \textcolor{black}{Kappa} \\
\midrule
EEGNet    & 0.2587$\pm$0.0010 & 0.1349$\pm$0.0012 & \underline{0.9430$\pm$0.0036} & \underline{0.9240$\pm$0.0048} & 0.2699$\pm$0.0023 & 0.0874$\pm$0.0029 & 0.5480$\pm$0.0224 & 0.0738$\pm$0.0353 \\
Conformer & 0.3209$\pm$0.0012 & 0.2074$\pm$0.0014 & 0.8981$\pm$0.0045 & 0.8641$\pm$0.0060 & 0.4170$\pm$0.0019 & 0.2713$\pm$0.0024 & 0.5984$\pm$0.0185 & 0.1552$\pm$0.0307 \\
TScep.    & 0.3300$\pm$0.0011 & 0.2180$\pm$0.0013 & 0.8026$\pm$0.0043 & 0.7368$\pm$0.0057 & \underline{0.5314$\pm$0.0024} & \underline{0.4143$\pm$0.0030} & \underline{0.6480$\pm$0.0177} & \underline{0.2397$\pm$0.0310} \\
STTran.   & 0.1867$\pm$0.0011 & 0.0511$\pm$0.0013 & 0.9398$\pm$0.0046 & 0.9197$\pm$0.0061 & 0.4266$\pm$0.0019 & 0.2833$\pm$0.0024 & 0.6335$\pm$0.0179 & 0.2146$\pm$0.0308 \\
BIOT      & 0.3096$\pm$0.0011 & 0.1943$\pm$0.0013 & 0.7936$\pm$0.0040 & 0.7248$\pm$0.0053 & 0.2920$\pm$0.0024 & 0.1150$\pm$0.0030 & 0.5857$\pm$0.0222 & 0.1343$\pm$0.0363 \\
EEGPT     & 0.1809$\pm$0.0013 & 0.0443$\pm$0.0015 & 0.9301$\pm$0.0047 & 0.9068$\pm$0.0063 & 0.2364$\pm$0.0022 & 0.0455$\pm$0.0028 & 0.4925$\pm$0.0192 & -0.0112$\pm$0.0286 \\
LaBraM    & 0.3244$\pm$0.0011 & 0.2115$\pm$0.0013 & 0.8698$\pm$0.0041 & 0.8264$\pm$0.0055 & 0.4819$\pm$0.0024 & 0.3524$\pm$0.0030 & 0.5650$\pm$0.0181 & 0.1008$\pm$0.0290 \\
CBraMod   & \underline{0.3311$\pm$0.0010} & \underline{0.2193$\pm$0.0012} & 0.9205$\pm$0.0039 & 0.8940$\pm$0.0052 & 0.4280$\pm$0.0022 & 0.2850$\pm$0.0028 & 0.5746$\pm$0.0200 & 0.1162$\pm$0.0324 \\
\midrule
\rowcolor{blue!8}
LEAF      & \textbf{0.3356$\pm$0.0012} & \textbf{0.2246$\pm$0.0014} & \textbf{0.9462$\pm$0.0043} & \textbf{0.9283$\pm$0.0057} & \textbf{0.5453$\pm$0.0023} & \textbf{0.4316$\pm$0.0029} & \textbf{0.6493$\pm$0.0211} & \textbf{0.2420$\pm$0.0370} \\
\bottomrule
\end{tabular}
\end{adjustbox}
\end{subtable}

\end{table}

We separately evaluate the fixed instruction-tuned model using cosine similarity to textual class prototypes, as described in Section~2.3. Figure~\ref{fig:ablation_summary}(D) compares a no-instruction baseline, a task-only instruction, and an instruction containing both the task and candidate classes. The no-instruction baseline uses a generic default string without task or class information. Mean balanced accuracy increases from 47.38\% to 48.67\% and 50.04\%, respectively. Including task and class information yields gains of 2.66 percentage points over the no-instruction baseline and 1.37 points over task-only conditioning. These results indicate that richer instructions improve direct decoding on held-out examples of the instruction-tuned tasks, without further parameter updates.

\begin{figure*}[!t]
    \centering
    \includegraphics[width=1.0\textwidth]{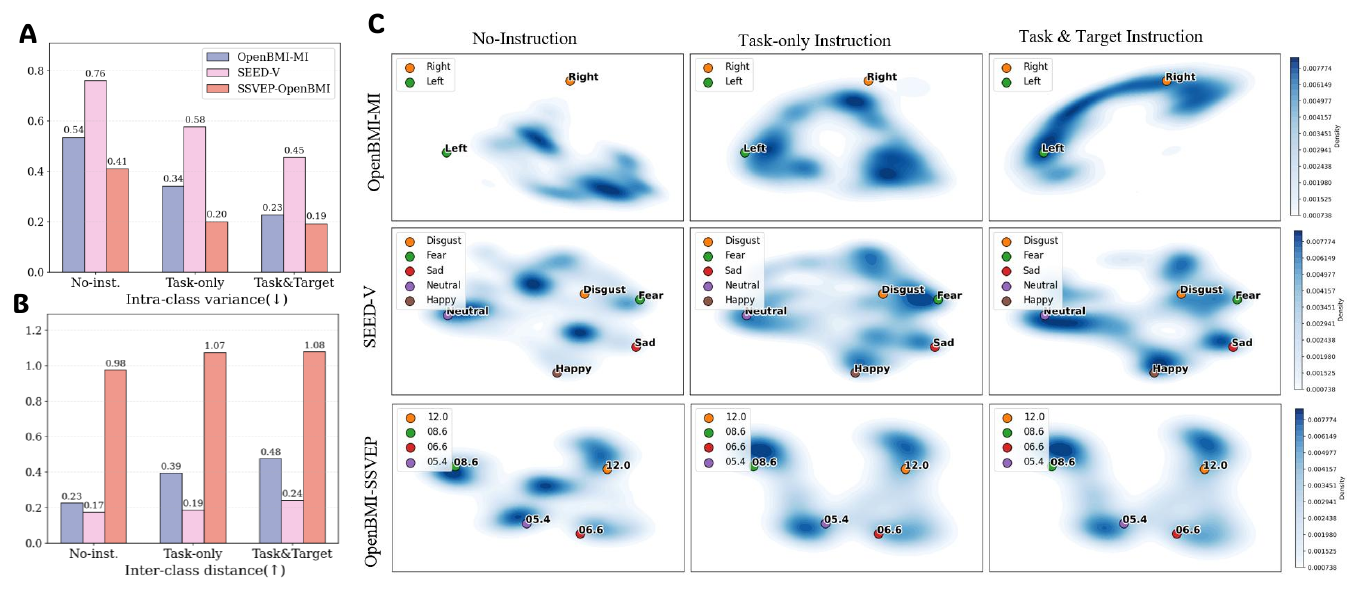}
    \caption{Effect of instruction conditioning on EEG representations: (A) intra-class variance, (B) inter-class distance, and (C) KDE-based embedding distributions with textual class prototypes.}
    \label{fig:feature1}
    \vspace{-8pt} 
\end{figure*}

\subsection{Analysis of Instruction-Guided EEG Representations}
\label{sec:instruction_analysis}
Figure~\ref{fig:feature1} shows a UMAP~\cite{mcinnes2018umap} projection of EEG embeddings with KDE sample densities and fixed textual class prototypes projected through the same mapping. Without instructions, embeddings show substantial class overlap; instruction conditioning produces more separated clusters with better alignment to their textual prototypes.

We quantify class compactness and separability using intra-class variance and inter-class distance. For class $c$ with embeddings $\{z_i\}_{i\in C_c}$ and mean $\mu_c$:
\begin{equation}
\text{Intra} = \frac{1}{K}\sum_{c=1}^K \frac{1}{|C_c|} \sum_{i\in C_c} \|z_i - \mu_c\|_2^2, \quad
\text{Inter} = \frac{2}{K(K-1)} \sum_{c<d} \|\mu_c - \mu_d\|_2,
\end{equation}
where $K$ is the number of classes. Figure~\ref{fig:feature1} shows lower intra-class variance and larger inter-class distances with instruction conditioning. Together, these observations suggest that task and candidate-class information helps organize EEG representations into more compact, class-discriminative groups within the evaluated tasks.

\subsection{Saliency Topographies}
\label{sec:saliency}

\begin{figure}[H]
\centerline{\includegraphics[width=0.95\textwidth]{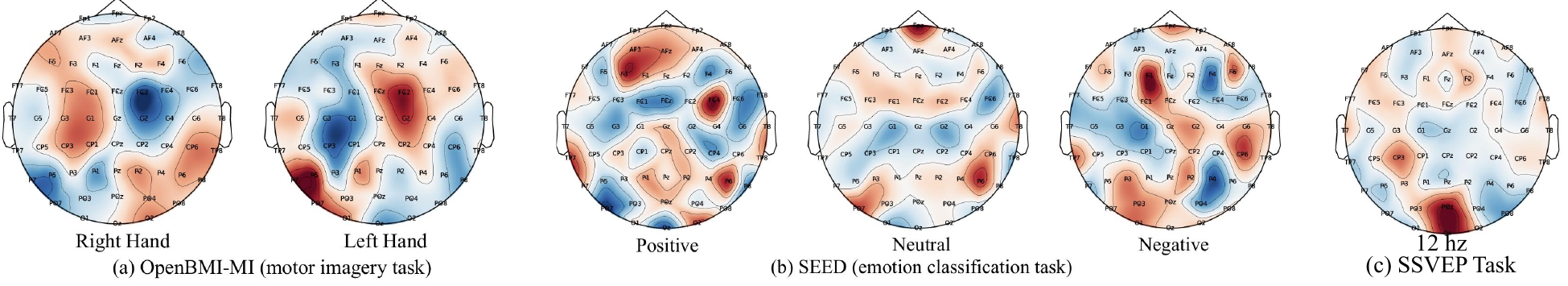}}
\caption{Topography visualization on downstream datasets, illustrating the spatial EEG activation patterns captured by LEAF across different BCI task categories.}
\label{fig:saliency}
\end{figure}

Figure~\ref{fig:saliency} presents gradient-based saliency topographies across representative downstream datasets. These maps highlight task-dependent spatial patterns, including motor cortex regions for MI, frontal/temporal regions for emotion recognition, and occipital regions for SSVEP.

\subsection{Q-Former Query Evolution}
\begin{figure*}[!htb]
    \centering
    \includegraphics[width=\textwidth]{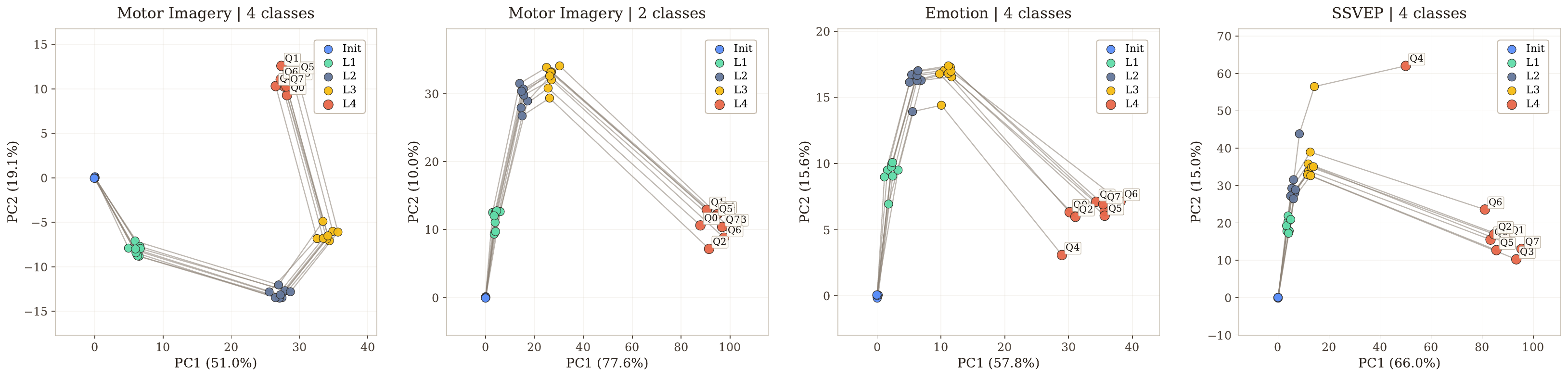}
    \caption{PCA visualization of Q-Former query evolution on four representative EEG decoding tasks. Queries diverge from their shared initialization and gradually form task-specific configurations across layers.}
    \label{fig:vis_q}
\end{figure*}

Figure~\ref{fig:vis_q} visualizes the layer-wise evolution of the IQF learnable queries using PCA on four representative EEG decoding tasks: 4-class motor imagery, 2-class motor imagery, 4-class emotion recognition, and 4-class SSVEP. Starting from a shared initialization, the queries progressively diverge across Q-Former layers and organize into task-specific configurations. This behavior suggests that the query set gradually specializes to capture task-relevant semantic structure, explaining why strong performance can be achieved with only a small number of learnable queries.

\subsection{Ablation Study}

\begin{figure}[H]
    \centering
    \includegraphics[width=1\textwidth]{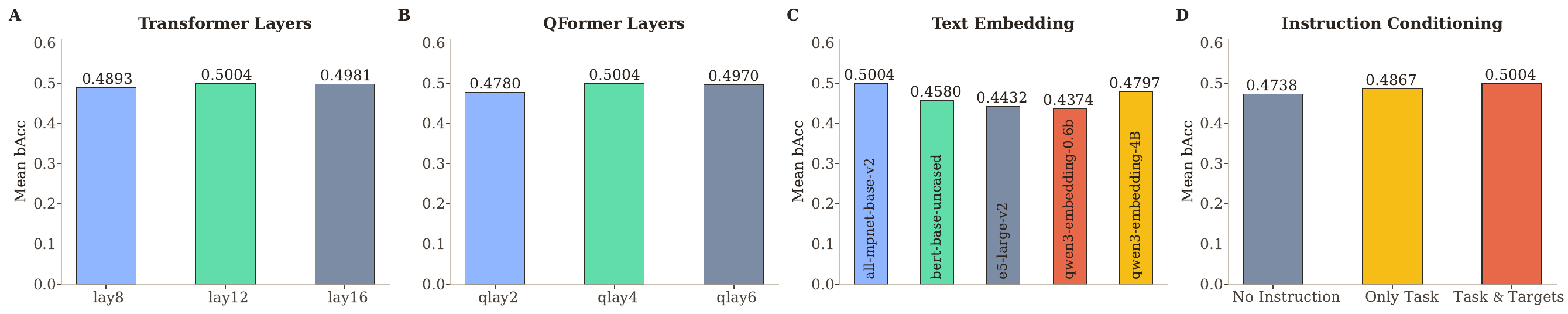}
    \caption{LEAF ablations: (A) EEG encoder depth, (B) Q-Former depth, (C) text encoder, and (D) instruction conditioning. Values are mean balanced accuracy.}
    \label{fig:ablation_summary}
\end{figure}
\vspace{-7pt} 

We examine model depth by varying the EEG encoder from 8 to 16 layers and the Q-Former from 2 to 6 layers (Figure~\ref{fig:ablation_summary}A--B). The highest mean balanced accuracy is obtained with 12 EEG encoder layers and 4 Q-Former layers; further increases provide no improvement. These results support the selected depths within the tested configurations, although the small numerical differences do not establish a statistically significant advantage.

We also examine how the choice of frozen text encoder and instruction content affects decoding (Figure~\ref{fig:ablation_summary}C--D). Among the text encoders, \texttt{all-mpnet-base-v2} achieves 50.04\% mean balanced accuracy, compared with 47.97\% for Qwen3-Embedding-4B, the next-best encoder tested. During direct inference, mean balanced accuracy increases from 47.38\% without task or class information to 48.67\% with task information and 50.04\% with both task and candidate-class information, supporting richer instruction conditioning within the evaluated tasks. Additional ablations of STR masking, query count, and cross-modal interaction are reported in Appendix~\ref{appendix:ablations}.

\section{Conclusion}
LEAF combines spectral-temporal reconstruction and an instruction-conditioned Q-Former to learn language-guided EEG representations. Supervised encoder finetuning achieves the highest average balanced accuracy among the evaluated models across 16 downstream datasets, while direct inference uses textual class prototypes without further parameter updates. These results support natural language as a source of semantic guidance for EEG foundation models.

\section*{\textcolor{black}{Funding and Competing Interests}}
\textcolor{black}{This research is supported by the Ministry of Education, Singapore, under its Academic Research Fund Tier 2 (Grant No. MOE-T2EP20124-0001). The authors declare no competing interests.}

\bibliographystyle{plainnat}
\bibliography{mybib}

@article{chen2023large,
  title={A large finer-grained affective computing EEG dataset},
  author={Chen, Jingjing and Wang, Xiaobin and Huang, Chen and Hu, Xin and Shen, Xinke and Zhang, Dan},
  journal={Scientific Data},
  volume={10},
  number={1},
  pages={740},
  year={2023},
  publisher={Nature Publishing Group UK London}
}

@inproceedings{chen2020simple,
  title={A simple framework for contrastive learning of visual representations},
  author={Chen, Ting and Kornblith, Simon and Norouzi, Mohammad and Hinton, Geoffrey},
  booktitle={International conference on machine learning},
  pages={1597--1607},
  year={2020},
  organization={PmLR}
}

@article{cho2017eeg,
  title={EEG datasets for motor imagery brain--computer interface},
  author={Cho, Hohyun and Ahn, Minkyu and Ahn, Sangtae and Kwon, Moonyoung and Jun, Sung Chan},
  journal={GigaScience},
  volume={6},
  number={7},
  pages={gix034},
  year={2017},
  publisher={Oxford University Press}
}

@inproceedings{devlin2019bert,
  title={Bert: Pre-training of deep bidirectional transformers for language understanding},
  author={Devlin, Jacob and Chang, Ming-Wei and Lee, Kenton and Toutanova, Kristina},
  booktitle={Proceedings of the 2019 conference of the North American chapter of the association for computational linguistics: human language technologies, volume 1 (long and short papers)},
  pages={4171--4186},
  year={2019}
}

@article{ding2022tsception,
  title={TSception: Capturing temporal dynamics and spatial asymmetry from EEG for emotion recognition},
  author={Ding, Yi and Robinson, Neethu and Zhang, Su and Zeng, Qiuhao and Guan, Cuntai},
  journal={IEEE Transactions on affective computing},
  volume={14},
  number={3},
  pages={2238--2250},
  year={2022},
  publisher={IEEE}
}

@inproceedings{duan2013differential,
  title={Differential entropy feature for EEG-based emotion classification},
  author={Duan, Ruo-Nan and Zhu, Jia-Yi and Lu, Bao-Liang},
  booktitle={2013 6th international IEEE/EMBS conference on neural engineering (NER)},
  pages={81--84},
  year={2013},
  organization={IEEE}
}

@article{edelman2024non,
  title={Non-invasive brain-computer interfaces: state of the art and trends},
  author={Edelman, Bradley J and Zhang, Shuailei and Schalk, Gerwin and Brunner, Peter and M{\"u}ller-Putz, Gernot and Guan, Cuntai and He, Bin},
  journal={IEEE reviews in biomedical engineering},
  volume={18},
  pages={26--49},
  year={2024},
  publisher={IEEE}
}

@inproceedings{he2020momentum,
  title={Momentum contrast for unsupervised visual representation learning},
  author={He, Kaiming and Fan, Haoqi and Wu, Yuxin and Xie, Saining and Girshick, Ross},
  booktitle={Proceedings of the IEEE/CVF conference on computer vision and pattern recognition},
  pages={9729--9738},
  year={2020}
}

@inproceedings{he2022masked,
  title={Masked autoencoders are scalable vision learners},
  author={He, Kaiming and Chen, Xinlei and Xie, Saining and Li, Yanghao and Doll{\'a}r, Piotr and Girshick, Ross},
  booktitle={Proceedings of the IEEE/CVF conference on computer vision and pattern recognition},
  pages={16000--16009},
  year={2022}
}

@article{jeong20222020,
  title={2020 International brain--computer interface competition: A review},
  author={Jeong, Ji-Hoon and Cho, Jeong-Hyun and Lee, Young-Eun and Lee, Seo-Hyun and Shin, Gi-Hwan and Kweon, Young-Seok and Mill{\'a}n, Jos{\'e} del R and M{\"u}ller, Klaus-Robert and Lee, Seong-Whan},
  journal={Frontiers in human neuroscience},
  volume={16},
  pages={898300},
  year={2022},
  publisher={Frontiers Media SA}
}

@article{jiang2026decoding,
  title={Decoding Covert Speech from EEG by Functional Areas Spatio-Temporal Transformer},
  author={Jiang, Muyun and Zhang, Wei and Ding, Yi and Teo, Kok Ann Colin and Fong, LaiGuan and Zhang, Shuailei and Guo, Zhiwei and Liu, Chenyu and Bhuvanakantham, Raghavan and Sim, Wei Khang Jeremy and others},
  journal={IEEE Journal of Biomedical and Health Informatics},
  year={2026},
  publisher={IEEE}
}

@article{jiang2024seed,
  title={SEED-VII: A multimodal dataset of six basic emotions with continuous labels for emotion recognition},
  author={Jiang, Wei-Bang and Liu, Xuan-Hao and Zheng, Wei-Long and Lu, Bao-Liang},
  journal={IEEE Transactions on Affective Computing},
  volume={16},
  number={2},
  pages={969--985},
  year={2024},
  publisher={IEEE}
}

@article{jiang2024neurolm,
  title={NeuroLM: A universal multi-task foundation model for bridging the gap between language and EEG signals},
  author={Jiang, Wei-Bang and Wang, Yansen and Lu, Bao-Liang and Li, Dongsheng},
  journal={arXiv preprint arXiv:2409.00101},
  year={2024}
}

@article{jiang2024large,
  title={Large brain model for learning generic representations with tremendous EEG data in BCI},
  author={Jiang, Wei-Bang and Zhao, Li-Ming and Lu, Bao-Liang},
  journal={arXiv preprint arXiv:2405.18765},
  year={2024}
}

@article{jiang2025towards,
  title={Towards robust multimodal physiological foundation models: Handling arbitrary missing modalities},
  author={Jiang, Wei-Bang and Fu, Xi and Ding, Yi and Guan, Cuntai},
  journal={arXiv preprint arXiv:2504.19596},
  year={2025}
}

@article{lawhern2018eegnet,
  title={EEGNet: a compact convolutional neural network for EEG-based brain--computer interfaces},
  author={Lawhern, Vernon J and Solon, Amelia J and Waytowich, Nicholas R and Gordon, Stephen M and Hung, Chou P and Lance, Brent J},
  journal={Journal of neural engineering},
  volume={15},
  number={5},
  pages={056013},
  year={2018},
  publisher={iOP Publishing}
}

@article{ouahidi2025reve,
  title={REVE: A Foundation Model for EEG--Adapting to Any Setup with Large-Scale Pretraining on 25,000 Subjects},
  author={Ouahidi, Yassine El and Lys, Jonathan and Th{\"o}lke, Philipp and Farrugia, Nicolas and Pasdeloup, Bastien and Gripon, Vincent and Jerbi, Karim and Lioi, Giulia},
  journal={arXiv preprint arXiv:2510.21585},
  year={2025}
}

@article{lee2019eeg,
  title={EEG dataset and OpenBMI toolbox for three BCI paradigms: An investigation into BCI illiteracy},
  author={Lee, Min-Ho and Kwon, O-Yeon and Kim, Yong-Jeong and Kim, Hong-Kyung and Lee, Young-Eun and Williamson, John and Fazli, Siamac and Lee, Seong-Whan},
  journal={GigaScience},
  volume={8},
  number={5},
  pages={giz002},
  year={2019},
  publisher={Oxford University Press}
}

@article{li2022eeg,
  title={EEG based emotion recognition: A tutorial and review},
  author={Li, Xiang and Zhang, Yazhou and Tiwari, Prayag and Song, Dawei and Hu, Bin and Yang, Meihong and Zhao, Zhigang and Kumar, Neeraj and Marttinen, Pekka},
  journal={ACM Computing Surveys},
  volume={55},
  number={4},
  pages={1--57},
  year={2022},
  publisher={ACM New York, NY}
}

@article{li2025investigating,
  title={Investigating the effects of sleep conditions on emotion responses with eeg signals and eye movements},
  author={Li, Ziyi and Tao, Le-Yan and Ma, Rui-Xiao and Zheng, Wei-Long and Lu, Bao-Liang},
  journal={IEEE Transactions on Affective Computing},
  year={2025},
  publisher={IEEE}
}

@article{liu2021comparing,
  title={Comparing recognition performance and robustness of multimodal deep learning models for multimodal emotion recognition},
  author={Liu, Wei and Qiu, Jie-Lin and Zheng, Wei-Long and Lu, Bao-Liang},
  journal={IEEE Transactions on Cognitive and Developmental Systems},
  volume={14},
  number={2},
  pages={715--729},
  year={2021},
  publisher={IEEE}
}

@article{liu2022identifying,
  title={Identifying similarities and differences in emotion recognition with EEG and eye movements among Chinese, German, and French People},
  author={Liu, Wei and Zheng, Wei-Long and Li, Ziyi and Wu, Si-Yuan and Gan, Lu and Lu, Bao-Liang},
  journal={Journal of Neural Engineering},
  volume={19},
  number={2},
  pages={026012},
  year={2022},
  publisher={IOP Publishing}
}

@inproceedings{ma2025codebrain,
  title={Codebrain: Bridging decoupled tokenizer and multi-scale architecture for eeg foundation model},
  author={Ma, Jingying and Wu, Feng and Lin, Qika and Xing, Yucheng and Liu, Chenyu and Jia, Ziyu and Feng, Mengling},
  booktitle={The Fourteenth International Conference on Learning Representations},
  year={2025}
}

@article{ma2022large,
  title={A large EEG dataset for studying cross-session variability in motor imagery brain-computer interface},
  author={Ma, Jun and Yang, Banghua and Qiu, Wenzheng and Li, Yunzhe and Gao, Shouwei and Xia, Xinxing},
  journal={Scientific Data},
  volume={9},
  number={1},
  pages={531},
  year={2022},
  publisher={Nature Publishing Group UK London}
}

@article{mcinnes2018umap,
  title={Umap: Uniform manifold approximation and projection for dimension reduction},
  author={McInnes, Leland and Healy, John and Melville, James},
  journal={arXiv preprint arXiv:1802.03426},
  year={2018}
}

@article{nieto2022thinking,
  title={Thinking out loud, an open-access EEG-based BCI dataset for inner speech recognition},
  author={Nieto, Nicol{\'a}s and Peterson, Victoria and Rufiner, Hugo Leonardo and Kamienkowski, Juan Esteban and Spies, Ruben},
  journal={Scientific data},
  volume={9},
  number={1},
  pages={52},
  year={2022},
  publisher={Nature Publishing Group UK London}
}

@article{perez2017learning,
  title={Learning visual reasoning without strong priors},
  author={Perez, Ethan and De Vries, Harm and Strub, Florian and Dumoulin, Vincent and Courville, Aaron},
  journal={arXiv preprint arXiv:1707.03017},
  year={2017}
}

@article{radford2019language,
  title={Language models are unsupervised multitask learners},
  author={Radford, Alec and Wu, Jeffrey and Child, Rewon and Luan, David and Amodei, Dario and Sutskever, Ilya and others},
  journal={OpenAI blog},
  volume={1},
  number={8},
  pages={9},
  year={2019}
}

@article{schalk2004bci2000,
  title={BCI2000: a general-purpose brain-computer interface (BCI) system},
  author={Schalk, Gerwin and McFarland, Dennis J and Hinterberger, Thilo and Birbaumer, Niels and Wolpaw, Jonathan R},
  journal={IEEE Transactions on biomedical engineering},
  volume={51},
  number={6},
  pages={1034--1043},
  year={2004},
  publisher={IEEE}
}

@article{schirrmeister2017deep,
  title={Deep learning with convolutional neural networks for EEG decoding and visualization},
  author={Schirrmeister, Robin Tibor and Springenberg, Jost Tobias and Fiederer, Lukas Dominique Josef and Glasstetter, Martin and Eggensperger, Katharina and Tangermann, Michael and Hutter, Frank and Burgard, Wolfram and Ball, Tonio},
  journal={Human brain mapping},
  volume={38},
  number={11},
  pages={5391--5420},
  year={2017},
  publisher={Wiley Online Library}
}

@article{shin2016open,
  title={Open access dataset for EEG+ NIRS single-trial classification},
  author={Shin, Jaeyoung and von L{\"u}hmann, Alexander and Blankertz, Benjamin and Kim, Do-Won and Jeong, Jichai and Hwang, Han-Jeong and M{\"u}ller, Klaus-Robert},
  journal={IEEE Transactions on Neural Systems and Rehabilitation Engineering},
  volume={25},
  number={10},
  pages={1735--1745},
  year={2016},
  publisher={IEEE}
}

@article{song2021transformer,
  title={Transformer-based spatial-temporal feature learning for EEG decoding},
  author={Song, Yonghao and Jia, Xueyu and Yang, Lie and Xie, Longhan},
  journal={arXiv preprint arXiv:2106.11170},
  year={2021}
}

@article{song2022eeg,
  title={EEG conformer: Convolutional transformer for EEG decoding and visualization},
  author={Song, Yonghao and Zheng, Qingqing and Liu, Bingchuan and Gao, Xiaorong},
  journal={IEEE Transactions on Neural Systems and Rehabilitation Engineering},
  volume={31},
  pages={710--719},
  year={2022},
  publisher={IEEE}
}

@article{stieger2021mindfulness,
  title={Mindfulness improves brain--computer interface performance by increasing control over neural activity in the alpha band},
  author={Stieger, James R and Engel, Stephen and Jiang, Haiteng and Cline, Christopher C and Kreitzer, Mary Jo and He, Bin},
  journal={Cerebral Cortex},
  volume={31},
  number={1},
  pages={426--438},
  year={2021},
  publisher={Oxford University Press}
}

@article{tangermann2012review,
  title={Review of the BCI competition IV},
  author={Tangermann, Michael and M{\"u}ller, Klaus-Robert and Aertsen, Ad and Birbaumer, Niels and Braun, Christoph and Brunner, Clemens and Leeb, Robert and Mehring, Carsten and Miller, Kai J and M{\"u}ller-Putz, Gernot R and others},
  journal={Frontiers in neuroscience},
  volume={6},
  pages={55},
  year={2012},
  publisher={Frontiers Research Foundation}
}

@article{valle2024identification,
  title={Identification of perceived sentences using deep neural networks in EEG},
  author={Valle, Carlos and Mendez-Orellana, Carolina and Herff, Christian and Rodriguez-Fernandez, Maria},
  journal={Journal of neural engineering},
  volume={21},
  number={5},
  pages={056044},
  year={2024},
  publisher={IOP Publishing}
}

@article{wairagkar2021dynamics,
  title={Dynamics of long-range temporal correlations in broadband EEG during different motor execution and imagery tasks},
  author={Wairagkar, Maitreyee and Hayashi, Yoshikatsu and Nasuto, Slawomir J},
  journal={Frontiers in neuroscience},
  volume={15},
  pages={660032},
  year={2021},
  publisher={Frontiers Media SA}
}

@article{wang2024eegpt,
  title={Eegpt: Pretrained transformer for universal and reliable representation of eeg signals},
  author={Wang, Guagnyu and Liu, Wenchao and He, Yuhong and Xu, Cong and Ma, Lin and Li, Haifeng},
  journal={Advances in Neural Information Processing Systems},
  volume={37},
  pages={39249--39280},
  year={2024}
}

@article{wang2024cbramod,
  title={Cbramod: A criss-cross brain foundation model for eeg decoding},
  author={Wang, Jiquan and Zhao, Sha and Luo, Zhiling and Zhou, Yangxuan and Jiang, Haiteng and Li, Shijian and Li, Tao and Pan, Gang},
  journal={arXiv preprint arXiv:2412.07236},
  year={2024}
}

@article{yang2023biot,
  title={Biot: Biosignal transformer for cross-data learning in the wild},
  author={Yang, Chaoqi and Westover, M and Sun, Jimeng},
  journal={Advances in Neural Information Processing Systems},
  volume={36},
  pages={78240--78260},
  year={2023}
}

@article{zheng2018emotionmeter,
  title={Emotionmeter: A multimodal framework for recognizing human emotions},
  author={Zheng, Wei-Long and Liu, Wei and Lu, Yifei and Lu, Bao-Liang and Cichocki, Andrzej},
  journal={IEEE transactions on cybernetics},
  volume={49},
  number={3},
  pages={1110--1122},
  year={2018},
  publisher={IEEE}
}

@article{zyma2019electroencephalograms,
  title={Electroencephalograms during mental arithmetic task performance},
  author={Zyma, Igor and Tukaev, Sergii and Seleznov, Ivan and Kiyono, Ken and Popov, Anton and Chernykh, Mariia and Shpenkov, Oleksii},
  journal={Data},
  volume={4},
  number={1},
  pages={14},
  year={2019},
  publisher={MDPI}
}

@article{zeng2025wavemind,
  title={Wavemind: Towards a conversational eeg foundation model aligned to textual and visual modalities},
  author={Zeng, Ziyi and Cai, Zhenyang and Cai, Yixi and Wang, Xidong and Chen, Junying and Wang, Rongsheng and Liu, Yipeng and Cai, Siqi and Wang, Benyou and Zhang, Zhiguo and others},
  journal={arXiv preprint arXiv:2510.00032},
  year={2025}
}

\clearpage
\appendix

\makeatletter
\renewcommand{\@seccntformat}[1]{%
  \ifnum\pdfstrcmp{#1}{section}=0
    Appendix~\csname the#1\endcsname\quad
  \else
    \csname the#1\endcsname\quad
  \fi
}
\makeatother

\begingroup
\setlength{\parindent}{0pt}
{\LARGE\bfseries Appendices\par}
\vspace{1.2em}
{\small\textsc{Contents}\hfill\textsc{Page}\par}
\vspace{0.4em}
\hrule height 0.5pt
\vspace{0.35em}
\newcommand{\appendixcontentsline}[2]{%
  \addlinespace[8pt]
  \hyperref[#1]{\textbf{\ref*{#1}\hspace{0.6em}#2}} &
  \hyperref[#1]{\textbf{\pageref*{#1}}} \\
}
\newcommand{\appendixsubcontentsline}[2]{%
  \hspace{0.8em}\hyperref[#1]{\ref*{#1}\hspace{0.5em}#2} &
  \hyperref[#1]{\pageref*{#1}} \\
}
\renewcommand{\arraystretch}{1.15}
\setlength{\tabcolsep}{4pt}
\begin{tabularx}{\textwidth}{@{}X r@{}}
\appendixcontentsline{appendix:background}{Related Work}
\appendixcontentsline{appendix:experimental_details}{Datasets and Preprocessing}
\appendixsubcontentsline{appendix:pretrain_datasets}{Pre-training Datasets}
\appendixsubcontentsline{appendix:downstream_datasets}{Downstream Dataset Details}
\appendixsubcontentsline{appendix:montage_config}{Electrode Montage and Channel Mapping}
\appendixcontentsline{appendix:implementation}{Implementation Details}
\appendixsubcontentsline{appendix:parameter}{Model and Training Configuration}
\appendixsubcontentsline{appendix:baselines}{Baseline Implementation}
\appendixsubcontentsline{appendix:compute}{Compute Resources}
\appendixcontentsline{appendix:additional_experiments}{Additional Experiments and Analysis}
\appendixsubcontentsline{appendix:additional_results}{Additional Results and Visualizations}
\appendixsubcontentsline{appendix:ablations}{Ablation Studies}
\appendixsubcontentsline{appendix:robustness}{Robustness and Generalization}
\appendixcontentsline{appendix:discussion}{Discussion}
\appendixsubcontentsline{appendix:limitations}{Limitations and Future Work}
\appendixsubcontentsline{appendix:broader_impacts}{Broader Impacts}
\end{tabularx}
\par\vspace{0.8em}
\hrule height 0.5pt
\endgroup

\section{Related Work}
\label{appendix:background}
\label{appendix:related_work}
\paragraph{Self-supervised Pretraining.} Self-supervised pretraining has emerged as a powerful paradigm in representation learning, reducing the reliance on large amounts of annotated data while leveraging abundant unlabeled signals. Self-supervised methods design pretext tasks that encourage models to learn meaningful feature representations from the inherent structure of data. Early successes in natural language processing, such as BERT \cite{devlin2019bert} and GPT series\cite{radford2019language}, demonstrated that masked language modeling and next-word prediction can yield representations transferable to diverse downstream tasks. Similarly, in computer vision, contrastive learning approaches like SimCLR\cite{chen2020simple}, MoCo\cite{he2020momentum} and masked image modeling like MAE\cite{he2022masked} showed that pretraining on large-scale unlabeled images leads to robust and generalizable visual features.
\paragraph{EEG Foundation Models.} The concept of foundation models has recently expanded into the EEG domain, aiming to build large-scale pre-trained backbones that generalize across datasets, tasks, and clinical conditions. Several pioneering efforts have been proposed. BIOT \cite{yang2023biot} explored scalable transformer-based architectures for biomedical signals, positioning EEG as a central modality. EEGPT \cite{wang2024eegpt}, inspired by advances in language modeling, leveraged transformer pretraining strategies such as masked prediction and contrastive learning to enhance generalization across heterogeneous EEG datasets. LaBraM \cite{jiang2024large} introduced a large-brain-model framework, emphasizing cross-dataset pretraining to capture universal EEG representations. CBraMod \cite{wang2024cbramod} extended this idea by focusing on cross-brain modularity, enabling adaptation across diverse cognitive and motor tasks. More recently, REVE \cite{ouahidi2025reve} introduced a versatile representation learning framework with a 4D positional encoding scheme that handles signals of arbitrary length and electrode arrangement, pretrained on over 60{,}000 hours of EEG from 25{,}000 subjects. CodeBrain \cite{ma2025codebrain} proposed a two-stage design that couples a decoupled time--frequency tokenizer (TFDual-Tokenizer) with a multi-scale architecture combining structured global convolutions and sliding-window attention, aligning with the brain's small-world topology for scalable representation learning. PhysioOmni \cite{jiang2025towards} further expanded the scope to multi-physiological modalities, integrating EEG with signals such as ECG and EMG to learn cross-modal representations. Collectively, these efforts highlight the emerging trajectory of EEG-FMs: moving from task-specific networks toward unified, pre-trained architectures capable of powering downstream applications with minimal fine-tuning, and paving the way for general-purpose brain decoding systems.
\paragraph{Text-aligned EEG Models.} A complementary line of work seeks to bridge EEG with natural language, leveraging the semantic richness of pretrained language models to enable instruction-following, zero-shot generalization, and human-interpretable outputs. NeuroLM \cite{jiang2024neurolm} proposed a broader neural language model for neuroscience data, treating EEG as a foreign language tokenized into a shared vocabulary with text so that a single LLM backbone can perform multi-task EEG decoding via natural-language prompts. WaveMind \cite{zeng2025wavemind} introduced a conversational EEG foundation model that aligns EEG representations with both textual and visual modalities, enabling unified interpretation of diverse brain activities through dialogue-style interfaces. Closer to the speech--language interface, FAST~\cite{jiang2026decoding} decodes covert speech directly from EEG, aligning neural activity with textual targets by exploiting the spatial organization of functional cortical regions together with temporal dynamics. These approaches signal a paradigm shift from purely signal-level pretraining toward language-aligned EEG foundation models that couple neural decoding with high-level semantic reasoning.

\section{Datasets and Preprocessing}
\label{appendix:experimental_details}
\subsection{Pre-training Datasets}
\label{appendix:pretrain_datasets}
\textcolor{black}{\textbf{Shared preprocessing.} The six datasets contain 325,046 segments at 200\,Hz, totaling 885.71 stored hours before loader padding, including training, validation, and preprocessing padding. All use 0.1--70\,Hz band-pass filtering, 0.1st--99.9th percentile clipping, median/IQR normalization, and mapping to 65 channels (Appendix~\ref{appendix:montage_config}). Normalization pools each Stieger session, SEED input recording, LargeSpanish subject--task pair, or ThinkOutLoud subject. Short segments are zero-padded to ten seconds, with random left/right allocation during training and right padding during validation. Labels are unused.}

\textcolor{black}{\textbf{Shared validation protocol.} Every tenth segment across alphabetically sorted files is reserved for validation, yielding 292,542 training and 32,504 validation segments before incomplete batches are dropped. This segment-level split does not hold out subjects or sessions and has no separate test set. All counts below follow this protocol.}

\subsubsection{Stieger2021 Motor Imagery Dataset}
\label{appendix:pretrain_stieger}
\textcolor{black}{The Stieger2021 recordings~\cite{stieger2021mindfulness} provide 62-channel EEG trials for motor imagery. We truncate or zero-pad these trials to ten seconds, resample them from 1,000 to 200\,Hz, and apply filtering. Additional 0.5th--99.5th percentile clipping precedes shared normalization; mapped values are clipped to $[-15,15]$. Six storage parts contain 254,250 segments (706.25 stored hours).}

\subsubsection{SEED French Emotion Dataset}
\label{appendix:pretrain_seed_fra}
\textcolor{black}{The French subset of SEED~\cite{liu2022identifying} provides emotion-related EEG recordings with 62 channels at 200\,Hz. Values are divided by 1,000 and filtered before non-overlapping ten-second segmentation and shared normalization; incomplete tails are discarded. The array contains 9,120 segments (25.33 hours).}

\subsubsection{SEED German Emotion Dataset}
\label{appendix:pretrain_seed_ger}
\textcolor{black}{For the German subset of SEED~\cite{liu2022identifying}, we follow the same preprocessing procedure as for SEED-FRA, yielding 11,309 ten-second segments (31.41 hours).}

\subsubsection{SEED Sleep Deprivation Dataset}
\label{appendix:pretrain_seed_sd}
\textcolor{black}{The SEED-SD dataset~\cite{li2025investigating} provides EEG recordings of emotional responses under sleep deprivation, recovery, and normal sleep. Its local \texttt{Sleep2} export uses the same preprocessing as SEED-FRA, yielding 25,068 ten-second segments (69.63 hours).}

\subsubsection{LargeSpanish Sentence Perception and Silent Production Dataset}
\label{appendix:pretrain_large_spanish}
\textcolor{black}{The LargeSpanish recordings~\cite{valle2024identification} cover sentence perception and silent production. We retain 62 EEG channels after auxiliary-channel removal, filtering, and resampling from 1,000 to 200\,Hz. Epochs span $-1$ to $+6$ seconds around event onset. Both tasks are pooled after separate normalization, yielding 20,640 seven-second segments (40.13 hours before loader padding).}

\subsubsection{ThinkOutLoud EEG Dataset}
\label{appendix:pretrain_think_out_loud}
\textcolor{black}{For ThinkOutLoud~\cite{nieto2022thinking}, we use three recording sessions per subject, retaining 128 EEG channels. The recordings are resampled to 200\,Hz and receive 50\,Hz notch and band-pass filtering. Sessions are concatenated before non-overlapping ten-second segmentation, discarding incomplete tails; segments may cross session boundaries. All recorded conditions are retained, yielding 4,659 segments (12.94 hours).}

\subsection{Downstream Dataset Details}
\label{appendix:downstream_datasets}
\textcolor{black}{All downstream datasets use the same normalization and montage-mapping procedure. Amplitudes are clipped at the 0.1st and 99.9th percentiles, then centered by the median and divided by the interquartile range of the clipped data. These scalar statistics pool trials, channels, and time points within each subject; for the SEED series, they are computed separately for each subject's training, validation, and test portions. Held-out subjects are normalized using their own recordings. Channels are mapped to the 65-channel LEAF montage, averaging any source channels assigned to the same template electrode and interpolating missing electrodes with MNE. The SEED-series exporters additionally clip the normalized, mapped arrays to $[-10,10]$. Dataset-specific filtering, segmentation, and evaluation splits are described below.}

\subsubsection{BCI Competition IV Dataset 2a}
\label{appendix:dataset_bcic_iv2a}
\textcolor{black}{Dataset 2a from BCI Competition IV~\cite{tangermann2012review} provides EEG recordings from nine subjects across two sessions, covering four motor imagery classes: left hand, right hand, foot, and tongue. We retain 22 EEG channels at 250\,Hz and exclude EOG. After skipping the first three recording blocks, trials are extracted using onset markers, re-referenced to the channel average, and cropped to their first six seconds. Both sessions are pooled within each subject, band-pass filtered to 0.3--40\,Hz, and resampled to 200\,Hz. The 2--6-second interval is retained, yielding 800 samples per trial. Labels 1--4 are shifted to 0--3; artifact annotations are not used for trial rejection.}

\textcolor{black}{\textbf{Evaluation split.} Subjects A01--A07 provide training and validation data, and A08--A09 form the cross-subject test set. Both sessions remain within the same subject partition.}

\subsubsection{OpenBMI Motor Imagery Dataset}
\label{appendix:dataset_openbmi_mi}
\textcolor{black}{The motor imagery subset of OpenBMI~\cite{lee2019eeg} provides EEG recordings from 54 subjects across two sessions, covering left- and right-hand imagery. For each subject, we concatenate the supplied training and test epochs from both sessions, retaining the 62 channels listed in the preprocessing script. The epochs are resampled from 1,000 to 200\,Hz and then band-pass filtered to 0.3--40\,Hz. Original labels 1 and 2 are mapped to 0 and 1, corresponding to right and left imagery. The supplied trial boundaries are preserved, yielding four-second inputs.}

\textcolor{black}{\textbf{Evaluation split.} Subjects s01--s42 provide training and validation data, and s43--s54 form the cross-subject test set.}

\subsubsection{BCI Competition Upper-Limb Imagery Dataset}
\label{appendix:dataset_upperlimb}
\textcolor{black}{The upper-limb imagery recordings from the BCI Competition dataset~\cite{jeong20222020} cover three types of imagined grasp: cylindrical, spherical, and lumbrical. We concatenate the supplied training and validation recordings for each of the 15 subjects, retain 60 EEG channels, and convert the one-hot labels to class indices. The final sample of each supplied epoch is removed before resampling to 200\,Hz. The epochs are then band-pass filtered to 0.1--70\,Hz, and the portion after the first six seconds is retained, corresponding to the final four-second imagery interval of the ten-second trial.}

\textcolor{black}{\textbf{Evaluation split.} Subjects 01--11 provide training and validation data; subjects 12--15 form the cross-subject test set.}

\subsubsection{Shanghai University Motor Imagery Dataset}
\label{appendix:dataset_shu}
\textcolor{black}{The ShanghaiU motor imagery dataset~\cite{ma2022large} provides left- and right-hand imagery recordings from 25 subjects across five sessions. The five session files are concatenated within each subject. Channels A1 and A2 are removed from the 32-channel input, leaving 30 EEG channels. The supplied epochs are resampled from 250 to 200\,Hz and band-pass filtered to 0.3--40\,Hz, retaining their trial boundaries. The original labels are shifted from 1--2 to 0--1.}

\textcolor{black}{\textbf{Evaluation split.} Subjects S01--S20 provide training and validation data, and S21--S25 form the cross-subject test set.}

\subsubsection{High-Gamma Motor Decoding Dataset}
\label{appendix:dataset_highgamma}
\textcolor{black}{For the HighGamma dataset~\cite{schirrmeister2017deep}, we use the supplied EEG epochs from 14 subjects. The preprocessing first retains the left-hand, right-hand, and feet classes and excludes rest. Channels M1 and M2 are removed from the 128-channel montage, leaving 126 channels before montage unification. Epochs are resampled to 200\,Hz and band-pass filtered to 0.3--40\,Hz.}

\textcolor{black}{\textbf{Evaluation split.} When constructing the evaluation splits, the dataset loader additionally removes the feet class, yielding the two-class left-versus-right task reported here. Subjects S01--S10 provide training and validation data; S11--S14 form the cross-subject test set.}

\subsubsection{Cho2017 Motor Imagery Dataset}
\label{appendix:dataset_cho}
\textcolor{black}{The motor imagery recordings released by Cho et al.~\cite{cho2017eeg} cover left- and right-hand imagery. We use epochs from 49 subjects, excluding identifiers s32, s46, and s49 from the original 52-subject list. The input uses the 64 channel names specified in the script. Signal values are divided by 100, resampled to 200\,Hz using the source sampling-rate metadata, and band-pass filtered to 0.3--40\,Hz. Each epoch is extended by 200 samples by repeating its final sample, providing one second of edge padding.}

\textcolor{black}{\textbf{Evaluation split.} The first 40 retained subjects provide training and validation data. The nine test subjects are s42, s43, s44, s45, s47, s48, s50, s51, and s52; these identifiers differ from consecutive positions in the retained subject list.}

\subsubsection{PhysioNet Motor Movement and Imagery Dataset (PhysioNet-MI)}
\label{appendix:dataset_physionet}
\textcolor{black}{\textcolor{black}{For PhysioNet-MI~\cite{schalk2004bci2000}, we use 64-channel recordings from subjects S001--S109 and pool runs 3--14, covering motor execution and imagery. The two non-rest event codes define the binary classes denoted ``Left'' and ``Right'' in Table~\ref{tab:dataset_overview}.} Channels marked as bad are interpolated, signals are resampled to 200\,Hz, and a 60\,Hz notch filter and a 0.3--40\,Hz band-pass filter are applied. Four-second epochs are extracted from annotation events without baseline correction, with the endpoint sample removed to retain 800 samples.}

\textcolor{black}{\textbf{Evaluation split.} Subjects S001--S080 provide training and validation data, and S081--S109 form the test set.}

\subsubsection{Shin2017A Motor Imagery Dataset}
\label{appendix:dataset_shin}
\textcolor{black}{For the motor imagery recordings released by Shin et al.~\cite{shin2016open}, we consider a binary hand-imagery task using the version supplied with ocular artifacts. Subject S05 is unavailable and excluded, leaving 28 subjects from S01--S29. We use recording blocks 1, 3, and 5, remove the vertical and horizontal EOG channels, and retain 30 EEG channels. Signals are treated as sampled at 200\,Hz, band-pass filtered to 0.3--40\,Hz, and notch filtered at 50\,Hz. Event markers define epochs from 0 to 4 seconds without baseline correction; removing the first sample yields 800 samples per epoch. The three blocks are pooled within each subject.}

\textcolor{black}{\textbf{Evaluation split.} The first 22 retained subjects provide training and validation data, and S24--S29 form the six-subject test set.}

\subsubsection{SEED Emotion Recognition Dataset}
\label{appendix:dataset_seed}
\textcolor{black}{For the SEED emotion recognition dataset~\cite{duan2013differential}, we use 62-channel EEG recordings from 15 subjects across three sessions. Negative, neutral, and positive labels are mapped to 0, 1, and 2. For each subject and trial index, the corresponding recordings from all three sessions are concatenated in time before filtering and segmentation. The supplied arrays are treated as sampled at 200\,Hz, band-pass filtered to 0.1--70\,Hz, and notch filtered at 50\,Hz with a 2\,Hz notch width. The concatenated signal is divided into non-overlapping four-second windows, discarding any incomplete trailing window. Windows can span a session boundary because sessions are concatenated before segmentation.}

\textcolor{black}{\textbf{Evaluation split.} Trial indices 1--9, 10--12, and 13--15 define training, validation, and test partitions, respectively, within the same subjects.}

\subsubsection{SEED-IV Emotion Recognition Dataset}
\label{appendix:dataset_seed_iv}
\textcolor{black}{The SEED-IV dataset~\cite{zheng2018emotionmeter} provides recordings from 15 subjects across three sessions, covering four emotion classes: neutral, sad, fear, and happy. Each of the 24 trials per session is processed separately using the supplied 62-channel arrays at 200\,Hz. We apply a 0.1--70\,Hz band-pass filter and a 50\,Hz notch filter with a 2\,Hz width, then divide each trial into non-overlapping four-second windows and discard any incomplete remainder. Labels follow the session-specific trial-label sequences.}

\textcolor{black}{\textbf{Evaluation split.} Within every session, trials 1--16 are assigned to training, 17--20 to validation, and 21--24 to testing. The corresponding partitions are pooled across sessions and subjects, preserving separation at the trial level.}

\subsubsection{SEED-V Emotion Recognition Dataset}
\label{appendix:dataset_seed_v}
\textcolor{black}{For the SEED-V dataset~\cite{liu2021comparing}, we use recordings from subjects 1--16 across three sessions, covering five emotion classes: disgust, fear, sad, neutral, and happy. The continuous recordings are loaded from CNT files; VEO, HEO, M1, and M2 are removed, retaining 62 EEG channels. Signals are resampled to 200\,Hz, band-pass filtered to 0.1--70\,Hz, and notch filtered at 50\,Hz with a 2\,Hz width. Session-specific start and end timestamps delimit the 15 emotion trials. Each trial is divided into non-overlapping four-second windows, discarding incomplete trailing samples.}

\textcolor{black}{\textbf{Evaluation split.} In each session, trials 1--5, 6--10, and 11--15 form training, validation, and test partitions, respectively. The partitions are then pooled across sessions and subjects.}

\subsubsection{SEED-VII Emotion Recognition Dataset}
\label{appendix:dataset_seed_vii}
\textcolor{black}{The SEED-VII dataset~\cite{jiang2024seed} provides recordings from 20 subjects across four sessions, with 20 trials per session. The seven classes are happy, surprise, neutral, sad, disgust, fear, and anger. M1, M2, ECG, HEO, and VEO are removed, leaving 62 EEG channels. Recordings are resampled to 200\,Hz, band-pass filtered to 0.1--70\,Hz, and notch filtered at 50\,Hz with a 2\,Hz width. Trial intervals are determined from paired start/end triggers; the implementation reconstructs triggers from auxiliary timestamp files for subject 14's first session and subject 9's third session. Each interval is segmented into non-overlapping four-second windows, discarding incomplete remainders.}

\textcolor{black}{\textbf{Evaluation split.} Within each session, trials 1--10 provide training data, 11--15 validation data, and 16--20 test data.}

\subsubsection{FACED Emotion Recognition Dataset}
\label{appendix:dataset_faced}
\textcolor{black}{For the FACED emotion recognition dataset~\cite{chen2023large}, we use the supplied trial arrays from 123 subjects, indexed 000--122, covering 28 film trials and nine emotion classes. The A1 and A2 channels are removed from the 32-channel input, leaving 30 EEG channels. Trials are band-pass filtered to 0.1--70\,Hz at 250\,Hz and then resampled to 200\,Hz. Each trial is divided into non-overlapping ten-second windows, with the film's label repeated for each window and any incomplete remainder discarded.}

\textcolor{black}{\textbf{Evaluation split.} Subjects 000--099 provide training and validation data; subjects 100--122 form the cross-subject test set. Thus, the test partition contains 23 subjects.}

\subsubsection{OpenBMI Steady-State Visual Evoked Potential Dataset}
\label{appendix:dataset_openbmi_ssvep}
\textcolor{black}{The SSVEP subset of OpenBMI~\cite{lee2019eeg} provides recordings from the same 54 subjects across two sessions. The four textual frequency labels are 12.0, 8.6, 6.6, and 5.4\,Hz. For each subject, the supplied training and test epochs from both sessions are concatenated, retaining 62 EEG channels. The epochs are resampled from 1,000 to 200\,Hz and band-pass filtered to 0.3--40\,Hz. Original labels 1--4 are shifted to 0--3, and the supplied four-second trial boundaries are preserved.}

\textcolor{black}{\textbf{Evaluation split.} Subjects s01--s42 provide training and validation data, and s43--s54 form the cross-subject test set.}

\subsubsection{BCI Competition Covert Speech Dataset}
\label{appendix:dataset_speech}
\textcolor{black}{The covert-speech recordings from the BCI Competition dataset~\cite{jeong20222020} provide 64-channel EEG epochs from 15 subjects, covering five labels: hello, help-me, stop, thank-you, and yes. For each subject, the original training and validation arrays are concatenated in that order, and the one-hot labels are converted to class indices. Five samples are appended to each epoch by repeating its final sample. The dataset-specific script performs no additional temporal filtering or resampling before applying the shared preprocessing.}

\textcolor{black}{\textbf{Evaluation split.} The loader constructs a cross-trial split within each subject: the first 250 concatenated trials are used for training, trials 251--300 for validation, and the remaining trials from 301 onward for testing. Corresponding partitions are pooled across subjects.}

\subsubsection{Mental Workload Dataset}
\label{appendix:dataset_workload}
\textcolor{black}{For mental-workload evaluation, we use EEG recordings from 36 subjects, indexed 00--35, collected before and during an arithmetic task~\cite{zyma2019electroencephalograms}. These two conditions serve as the resting and workload classes. The ECG and A2--A1 channels are removed, retaining 19 EEG channels. Each continuous recording is resampled to 200\,Hz and band-pass filtered to 0.1--70\,Hz. The two conditions are segmented separately into non-overlapping four-second windows, with incomplete trailing samples discarded, and are assigned labels 0 and 1 before concatenation within each subject.}

\textcolor{black}{\textbf{Evaluation split.} Subjects 00--31 provide training and validation data, and subjects 32--35 form the cross-subject test set.}

\begingroup
\raggedbottom
\subsection{Electrode Montage and Channel Mapping}
\label{appendix:montage_config}
LEAF maps heterogeneous electrode layouts to a common 65-channel montage with a fixed channel order. A predefined mapping assigns 349 distinct source-channel names to the template electrodes, accommodating differences in electrode naming and placement across datasets. Figure~\ref{fig:LEAF_Montage} illustrates the target montage, and Table~\ref{tab:channel_mapping} lists the complete source-to-target assignments in the output channel order.

The mapping includes legacy naming conventions, such as T3$\rightarrow$T7, T4$\rightarrow$T8, T5$\rightarrow$P7, and T6$\rightarrow$P8, as well as predefined spatial groupings, such as \{FCz, FFCz\}$\rightarrow$FCz and \{Oz, Iz, OIz\}$\rightarrow$Oz. When several available source electrodes map to the same template electrode, their signals are averaged pointwise with equal weights. A single assigned source is retained unchanged. Only electrodes present in the recording contribute to the average. These groupings are fixed preprocessing choices and do not imply that the grouped electrode locations are identical.

\textcolor{black}{For example, the Fp1 group contains \{Fp1, Fp1h, AFp1h, AFp1, AFp3, AFp5, AFp5h, AFp7h\}. If a recording contains only Fp1 and AFp1 from this group, the output is $y_{\mathrm{Fp1}}(t)=[x_{\mathrm{Fp1}}(t)+x_{\mathrm{AFp1}}(t)]/2$. If only Fp1 is present, its signal is retained unchanged; if none of these eight channels is present, the target Fp1 is interpolated. More generally, for target electrode $e$, let $P_e$ be its configured source-channel group and $C$ the source channels present in the recording. With $A_e=P_e\cap C$, the mapped signal for each trial is}
\begingroup
\color{black}
\setlength{\abovedisplayskip}{8pt}
\setlength{\belowdisplayskip}{8pt}
\setlength{\abovedisplayshortskip}{4pt}
\setlength{\belowdisplayshortskip}{8pt}
\begin{equation}
y_e(t)=\frac{1}{|A_e|}\sum_{c\in A_e}x_c(t),\qquad |A_e|>0.
\end{equation}
\endgroup

\begin{table}[H]
\centering
\caption{Complete source-to-template channel mapping for LEAF.}
\label{tab:channel_mapping}
\begingroup
\color{black}
\fontsize{8}{9}\selectfont
\setlength{\tabcolsep}{3pt}
\renewcommand{\arraystretch}{1.0}
\begin{tabularx}{\textwidth}{@{}c l >{\raggedright\arraybackslash}X@{}}
\toprule
\textbf{Index} & \textbf{Target} & \textbf{Source channel names assigned to the target} \\
\midrule
1 & Fpz & Fpz, AFpz \\
2 & AFz & AFz, AFFz \\
3 & Fz & Fz \\
4 & FCz & FCz, FFCz \\
5 & Cz & Cz, FCCz \\
6 & CPz & CPz, CCPz \\
7 & Pz & Pz, CPPz \\
8 & POz & POz, PPOz, POOz \\
9 & Oz & Oz, Iz, OIz \\
10 & Fp1 & Fp1, Fp1h, AFp1h, AFp1, AFp3, AFp5, AFp5h, AFp7h \\
11 & Fp2 & Fp2, Fp2h, AFp2h, AFp2, AFp4, AFp6, AFp6h, AFp8h \\
12 & AF7 & AF5, AF7, AF7h, AFp7, AFF7h, AF9, AFp9h, AF9h, AFp9 \\
13 & AF3 & AF1, AF1h, AF3, AF3h, AFp3h, AFF3h, AF5h \\
14 & AF4 & AF2, AF2h, AF4, AF4h, AFp4h, AFF4h, AF6h \\
15 & AF8 & AF6, AF8, AF8h, AFp8, AFF8h, AF10, AFp10h, AF10h, AFp10 \\
16 & F7 & F7, F7h, AFF7, AFF9h, FFT7h, F9, F9h, AFF9 \\
17 & F5 & F5, F5h, AFF5, FFC5h, FFC5 \\
18 & F3 & F3, F3h, AFF3, FFC3, AFF5h \\
19 & F1 & F1, F1h, AFF1, FFC3h, AFF1h \\
20 & F2 & F2, F2h, AFF2, FFC4h, AFF2h \\
21 & F4 & F4, F4h, AFF4, FFC4, AFF6h \\
22 & F6 & F6, F6h, AFF6, FFC6h, FFC6 \\
23 & F8 & F8, F8h, AFF8, AFF10h, FFT8h, F10, F10h, AFF10 \\
24 & FT9 & FT9, FFT9h, FTT9, FTT9h, FT9h, FFT9 \\
25 & FT7 & FT7, FTT7h, FT7h, FFT7 \\
26 & FC5 & FC5, FC5h, FCC5 \\
27 & FC3 & FC3, FC3h, FCC3 \\
28 & FC1 & FC1, FFC1, FCC1, FFC1h, FC1h \\
29 & FC2 & FC2, FFC2, FCC2, FFC2h, FC2h \\
30 & FC4 & FC4, FC4h, FCC4 \\
31 & FC6 & FC6, FC6h, FCC6 \\
32 & FT8 & FT8, FTT8h, FT8h, FFT8 \\
33 & FT10 & FT10, FFT10, FTT10, FTT10h, FFT10h, FT10h \\
34 & T7 & T7, T3, T7h, T9, T9h, FTT7 \\
35 & C5 & C5, C5h, FCC5h, CCP5 \\
36 & C3 & C3, C3h, FCC3h, CCP3 \\
37 & C1 & C1, C1h, FCC1h, CCP1 \\
38 & C2 & C2, C2h, FCC2h, CCP2 \\
39 & C4 & C4, C4h, FCC4h, CCP4 \\
40 & C6 & C6, C6h, FCC6h, CCP6 \\
41 & T8 & T8, T4, T8h, T10, T10h, FTT8 \\
42 & TP9 & TP9, P9, P11, TTP9h, TTP9, TPP9 \\
43 & TP7 & TP7, TP7h, TPP7, TTP7, TTP7h, TPP9h, TP9h \\
44 & CP5 & CP5, CCP5h, CP5h, CPP5 \\
45 & CP3 & CP3, CP3h, CCP3h \\
46 & CP1 & CP1, CP1h, CCP1h \\
47 & CP2 & CP2, CP2h, CCP2h \\
48 & CP4 & CP4, CP4h, CCP4h \\
49 & CP6 & CP6, CCP6h, CP6h, CPP6 \\
50 & TP8 & TP8, TP8h, TPP8, TTP8, TTP8h, TPP10h, TP10h \\
51 & TP10 & TP10, P10, P12, TTP10, TTP10h, TPP10 \\
52 & P7 & T5, P7, P7h, TPP7h, PPO9h, P9h, PPO9 \\
53 & P5 & P5, P5h, CPP5h, PPO5, TPP5h \\
54 & P3 & P3, P3h, CPP3, PPO3, CPP3h, PPO3h \\
55 & P1 & P1, P1h, CPP1, CPP1h, PPO1 \\
56 & P2 & P2, P2h, CPP2, CPP2h, PPO2 \\
57 & P4 & P4, P4h, CPP4, PPO4, CPP4h, PPO4h \\
58 & P6 & P6, P6h, CPP6h, PPO6, TPP6h \\
59 & P8 & T6, P8, P8h, TPP8h, PPO10h, P10h, PPO10 \\
60 & PO7 & PO7, PO9, PPO7h, POO9h, PO9h, PO7h, PPO7, POO9, PO11, POO11h \\
61 & PO3 & PO1, PO1h, PPO1h, PO5, PO3, PO3h, POO3h, POO3, PPO5h, PO5h \\
62 & PO4 & PO2, PO2h, PPO2h, PO6, PO4, PO4h, POO4h, POO4, PPO6h, PO6h \\
63 & PO8 & PO8, PO10, PPO8h, POO10h, PO8h, PO10h, PPO8, POO10, PO12, POO12h \\
64 & O1 & O1, O1h, I1, I1h, OI1h, POO1, POO1h, POO7h, POO5h, POO7, POO5, OI1, CB2 \\
65 & O2 & O2, O2h, I2, I2h, OI2h, POO2, POO2h, POO6h, POO8h, POO6, POO8, OI2, CB1 \\
\bottomrule
\end{tabularx}
\endgroup
\end{table}

Template electrodes without an available source channel are spatially interpolated using MNE and the template electrode coordinates. Interpolation is unnecessary when all template electrodes have assigned source signals. Source-channel names outside the predefined mapping are handled during dataset-specific preprocessing. The resulting signals follow the same 65-channel order across datasets, giving inputs of shape $B\times65\times T$, where $B$ is the number of trials and $T$ is the number of time samples at 200\,Hz.

\begin{figure}[H]
\centering
\includegraphics[width=0.5\textwidth]{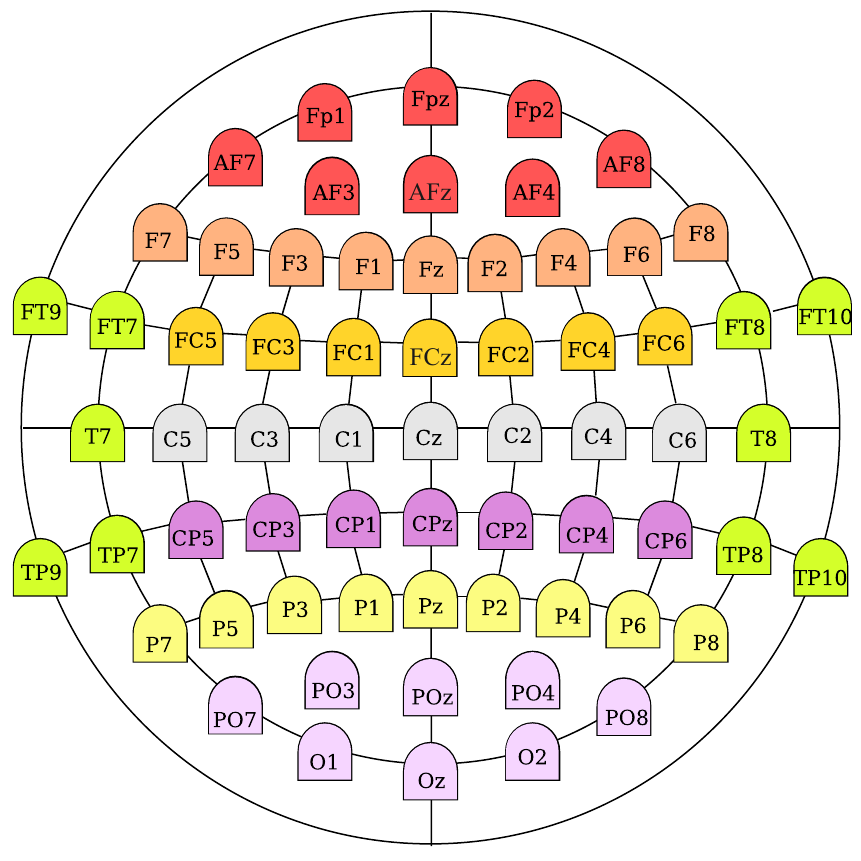}
\caption{Standardized 65-channel EEG montage used by LEAF. Inputs from different datasets are mapped to this common 10--10 electrode configuration by source-channel assignment and averaging, with spatial interpolation for missing electrodes.}
\label{fig:LEAF_Montage}
\end{figure}

\endgroup
\section{Implementation Details}
\label{appendix:implementation}
\subsection{Model and Training Configuration}
\label{appendix:parameter}
\textcolor{black}{Table~\ref{tab:all_hyperparameters} summarizes the MPNet-based architecture, and Table~\ref{tab:training_settings} lists the stage-specific training defaults in the supplied configuration and entry points.}

\begin{table}[htbp]
\centering
\caption{\textcolor{black}{LEAF architecture with the MPNet text encoder.}}
\label{tab:all_hyperparameters}
\begingroup
\color{black}
\footnotesize
\setlength{\tabcolsep}{4pt}
\renewcommand{\arraystretch}{1.08}
\begin{tabularx}{\textwidth}{@{}l l >{\raggedright\arraybackslash}X@{}}
\toprule
\textbf{Group} & \textbf{Parameter} & \textbf{Value} \\
\midrule
Input & Channels & 65 \\
 & Sampling rate & 200\,Hz \\
 & Window length / stride & 100 / 100 samples (0.5\,s; non-overlapping) \\
\midrule
Tokenizer & Convolutional filters & 64 \\
 & Temporal kernel / padding & $(1,40)$ / $(0,20)$ \\
 & Spatial kernel & $(65,1)$ \\
 & Max pooling & $(1,10)$ \\
 & Dropout & 0.5 \\
\midrule
EEG encoder & Branches & 2 \\
 & Transformer layers per branch & 12 \\
 & Token dimension & 256 \\
 & Attention heads & 8 \\
 & FFN expansion factor & 4 \\
 & Dropout & 0.1 \\
\midrule
STR masking & Temporal mask probability & 0.5 \\
 & Frequency mask & $[f,f+6)$\,Hz; integer $f$ sampled uniformly from 1--50\,Hz \\
\midrule
IQF and text & Queries / layers & 8 / 4 \\
 & Text encoder & \texttt{all-mpnet-base-v2} \\
 & Text embedding dimension & 768 \\
\midrule
Model size & EEG tower parameters & 20.230\,M (including tokenizer: 0.269\,M) \\
 & Q-Former parameters & 6.194\,M \\
 & Total parameters & 26.424\,M \\
\bottomrule
\end{tabularx}

\endgroup
\end{table}

\begin{table}[htbp]
\centering
\caption{LEAF stage-specific training defaults.}
\label{tab:training_settings}
\begingroup
\color{black}
\footnotesize
\setlength{\tabcolsep}{4pt}
\renewcommand{\arraystretch}{1.08}
\begin{tabularx}{\textwidth}{@{}l *{3}{>{\centering\arraybackslash}X}@{}}
\toprule
\textbf{Training setting} & \textbf{Pretraining} & \textbf{Instruction tuning} & \textbf{Finetuning} \\
\midrule
Peak LR: CNN (Fast) & $10^{-6}$ & $10^{-3}$ & $10^{-3}$ \\
Peak LR: Transformers (Slow) & $10^{-7}$ & $2\!\times\!10^{-4}$ & $10^{-4}$ \\
Warmup epochs & 1 & 1 & 1 \\
Total epochs & 10 & 100 & 100 \\
Global batch size & 512 & 512 & 128 \\
Weight decay & $10^{-2}$ & $10^{-3}$ & $10^{-3}$ \\
\bottomrule
\end{tabularx}

\endgroup
\end{table}

\textcolor{black}{All stages use AdamW ($\beta_1=0.9$, $\beta_2=0.999$), bf16 mixed precision, and linear warmup followed by cosine decay toward 0.1 times each group's peak learning rate. The fast group contains the tokenizer and, during finetuning, the classification head; all remaining parameters belong to the slow group. The two 512-sample batches follow the supplied training script (256 per GPU on two GPUs); finetuning uses the entry-point default of 128 on one GPU. Pretraining uses the entry-point default of 10 epochs. Command-line overrides can change these defaults. Parameter counts exclude the frozen text encoder and dataset-specific classification heads.}

\subsection{Baseline Implementation}
\label{appendix:baselines}
To ensure a fair comparison, all baseline models were re-trained or fine-tuned using their officially released implementations and recommended hyperparameters. For each dataset, all baselines used the same input trial duration as LEAF, with EEG resampled to 200\,Hz and mapped to 65 channels. Dataset-specific trial lengths followed the preprocessing described in Appendix~\ref{appendix:downstream_datasets}. For transformer-based foundation models, these trials were segmented into non-overlapping 200-sample windows as tokens; this window length specifies tokenization, not the total input trial duration. All models were trained for 100 epochs using the Adam optimizer with a learning rate of $1\times10^{-3}$ and cross-entropy loss. Pretrained checkpoints of LaBraM, EEGPT, and CBraMod were loaded and fine-tuned end-to-end from their official releases; for LaBraM and NeuroLM, we adopted the base-model variant. Because NeuroLM is a generative model without a straightforward task-specific fine-tuning protocol, a direct comparison in the task-specific finetuning setting is not applicable. We therefore compare LEAF with NeuroLM exclusively under the direct inference setting, where LEAF is evaluated after multi-task instruction tuning without any further fine-tuning.

\subsection{Compute Resources}
\label{appendix:compute}
All experiments were conducted on a single workstation equipped with 2$\times$NVIDIA RTX PRO 5000 GPUs (Blackwell architecture, 48\,GB VRAM each), an AMD Ryzen Threadripper PRO 5975WX CPU (32 cores / 64 threads), 512\,GB system memory, and local NVMe SSD storage, running Ubuntu 24.04 LTS. Pre-training took approximately 4 hours and instruction tuning took approximately 2 hours.

\section{Additional Experiments and Analysis}
\label{appendix:additional_experiments}
\subsection{Additional Results and Visualizations}
\label{appendix:additional_results}
This section complements the main results with a comparison of finetuning strategies and visualizations of the learned representations. UMAP projections illustrate the organization of EEG embeddings and textual prototypes, complementing the saliency analysis in Section~\ref{sec:saliency}.

\subsubsection{Finetuning with the Q-Former}
\label{appendix:qformer_finetuning}
\begin{figure}[H]
    \centering
    \includegraphics[width=\textwidth]{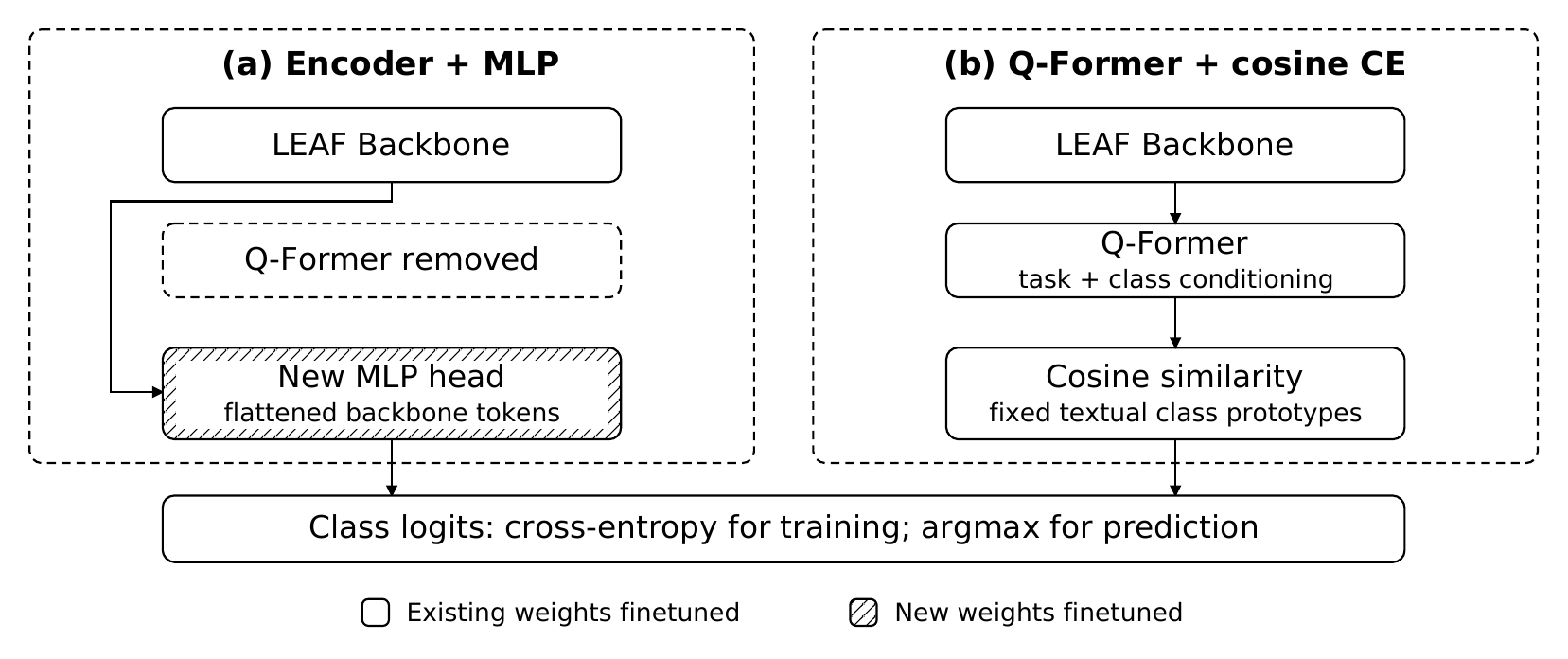}
    \caption{LEAF finetuning from a shared instruction-tuned checkpoint: (a) replace the Q-Former with an MLP head; (b) retain the encoder and Q-Former and apply cross-entropy to cosine similarities with fixed text prototypes, adding no trainable parameters.}
    \label{fig:finetuning_paths}
\end{figure}

Figure~\ref{fig:finetuning_paths} illustrates two supervised classification paths. The encoder-plus-MLP setting removes the Q-Former to facilitate baseline comparison under a conventional supervised classification setup. The complementary Q-Former + cosine CE setting evaluates LEAF's learned EEG--language alignment without adding trainable parameters: cross-entropy on cosine similarities to fixed text prototypes updates the existing encoder and Q-Former. Table~\ref{tab:finetuning_comparison} compares this LEAF-specific adaptation with the main-table LEAF results from Tables~\ref{tab:mean_std} and~\ref{tab:mean_std_other}.

\begin{table}[H]
\centering
\footnotesize
\caption{LEAF finetuning comparison for the two paths in Fig.~\ref{fig:finetuning_paths}. B.Acc and Kappa are mean $\pm$ standard deviation; the average row gives equal-weight means across 16 datasets.}
\label{tab:finetuning_comparison}
\setlength{\tabcolsep}{5pt}
\renewcommand{\arraystretch}{1.12}
\begin{tabular*}{0.94\textwidth}{@{\extracolsep{\fill}}rlcccc@{}}
\toprule
\multirow{2}{*}{ID} & \multirow{2}{*}{Dataset} & \multicolumn{2}{c}{\makecell{\textbf{(A) Encoder + MLP (Main)}\\New parameters: MLP head}} & \multicolumn{2}{c}{\makecell{\textbf{(B) Q-Former + cosine CE}\\New parameters: none}} \\
\cmidrule(lr){3-4}\cmidrule(lr){5-6}
& & B.Acc & Kappa & B.Acc & Kappa \\
\midrule
1 & BCIC-IV-2a & $0.6381 \pm 0.0120$ & $0.5175 \pm 0.0160$ & $0.6620 \pm 0.0136$ & $0.5494 \pm 0.0182$ \\
2 & OpenBMI-MI & $0.8134 \pm 0.0031$ & $0.6268 \pm 0.0062$ & $0.8017 \pm 0.0018$ & $0.6035 \pm 0.0036$ \\
3 & BCIC-Upperlimb & $0.5483 \pm 0.0038$ & $0.3225 \pm 0.0057$ & $0.4828 \pm 0.0038$ & $0.2242 \pm 0.0058$ \\
4 & SHU-MI & $0.6134 \pm 0.0008$ & $0.2268 \pm 0.0016$ & $0.5897 \pm 0.0053$ & $0.1792 \pm 0.0106$ \\
5 & HighGamma & $0.7982 \pm 0.0035$ & $0.5964 \pm 0.0070$ & $0.7899 \pm 0.0104$ & $0.5797 \pm 0.0208$ \\
6 & Cho2017 & $0.7908 \pm 0.0060$ & $0.5816 \pm 0.0120$ & $0.8030 \pm 0.0056$ & $0.6059 \pm 0.0111$ \\
7 & Shin2017A & $0.7256 \pm 0.0088$ & $0.4512 \pm 0.0176$ & $0.7509 \pm 0.0307$ & $0.5019 \pm 0.0614$ \\
8 & PhysioNet-MI & $0.6992 \pm 0.0035$ & $0.3984 \pm 0.0070$ & $0.7535 \pm 0.0025$ & $0.5070 \pm 0.0050$ \\
\midrule
9 & FACED & $0.5819 \pm 0.0023$ & $0.5293 \pm 0.0026$ & $0.4071 \pm 0.0040$ & $0.3292 \pm 0.0050$ \\
10 & SEED & $0.7011 \pm 0.0068$ & $0.5511 \pm 0.0102$ & $0.6870 \pm 0.0035$ & $0.5347 \pm 0.0045$ \\
11 & SEED-IV & $0.4630 \pm 0.0034$ & $0.2660 \pm 0.0044$ & $0.4511 \pm 0.0048$ & $0.2702 \pm 0.0056$ \\
12 & SEED-V & $0.4126 \pm 0.0023$ & $0.2640 \pm 0.0029$ & $0.4010 \pm 0.0076$ & $0.2556 \pm 0.0106$ \\
13 & SEED-VII & $0.3356 \pm 0.0012$ & $0.2246 \pm 0.0014$ & $0.3286 \pm 0.0032$ & $0.2175 \pm 0.0038$ \\
\midrule
14 & OpenBMI-SSVEP & $0.9462 \pm 0.0043$ & $0.9283 \pm 0.0057$ & $0.8971 \pm 0.0047$ & $0.8628 \pm 0.0063$ \\
15 & BCIC-Speech & $0.5453 \pm 0.0023$ & $0.4316 \pm 0.0029$ & $0.5338 \pm 0.0078$ & $0.4172 \pm 0.0098$ \\
16 & Mental Workload & $0.6493 \pm 0.0211$ & $0.2420 \pm 0.0370$ & $0.6296 \pm 0.0334$ & $0.3122 \pm 0.0921$ \\
\midrule
 & \textbf{Average} & 0.6414 & 0.4474 & 0.6231 & 0.4344 \\
\bottomrule
\end{tabular*}
\end{table}

\subsubsection{UMAP Embedding Visualizations}
\label{appendix:umap}
\begin{figure*}[h]
    \centering
    \includegraphics[width=0.95\textwidth]{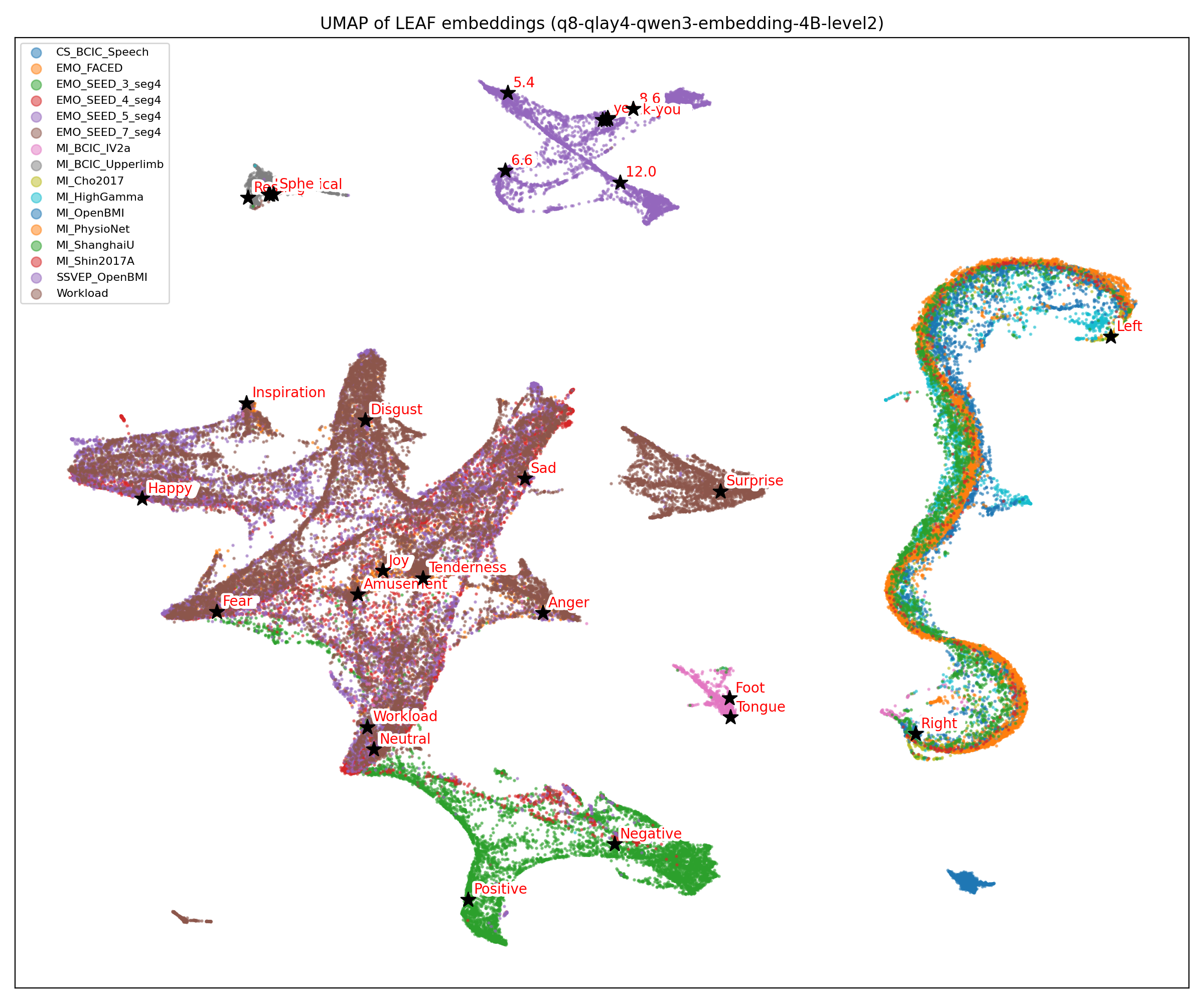}
    \caption{\textcolor{black}{UMAP projection of EEG embeddings and textual class prototypes across downstream datasets under task-and-target instruction. Colors indicate datasets; stars indicate textual class prototypes.}}
    \label{fig:umap_global}
\end{figure*}

\begin{figure*}[h]
    \centering
    \includegraphics[width=0.95\textwidth]{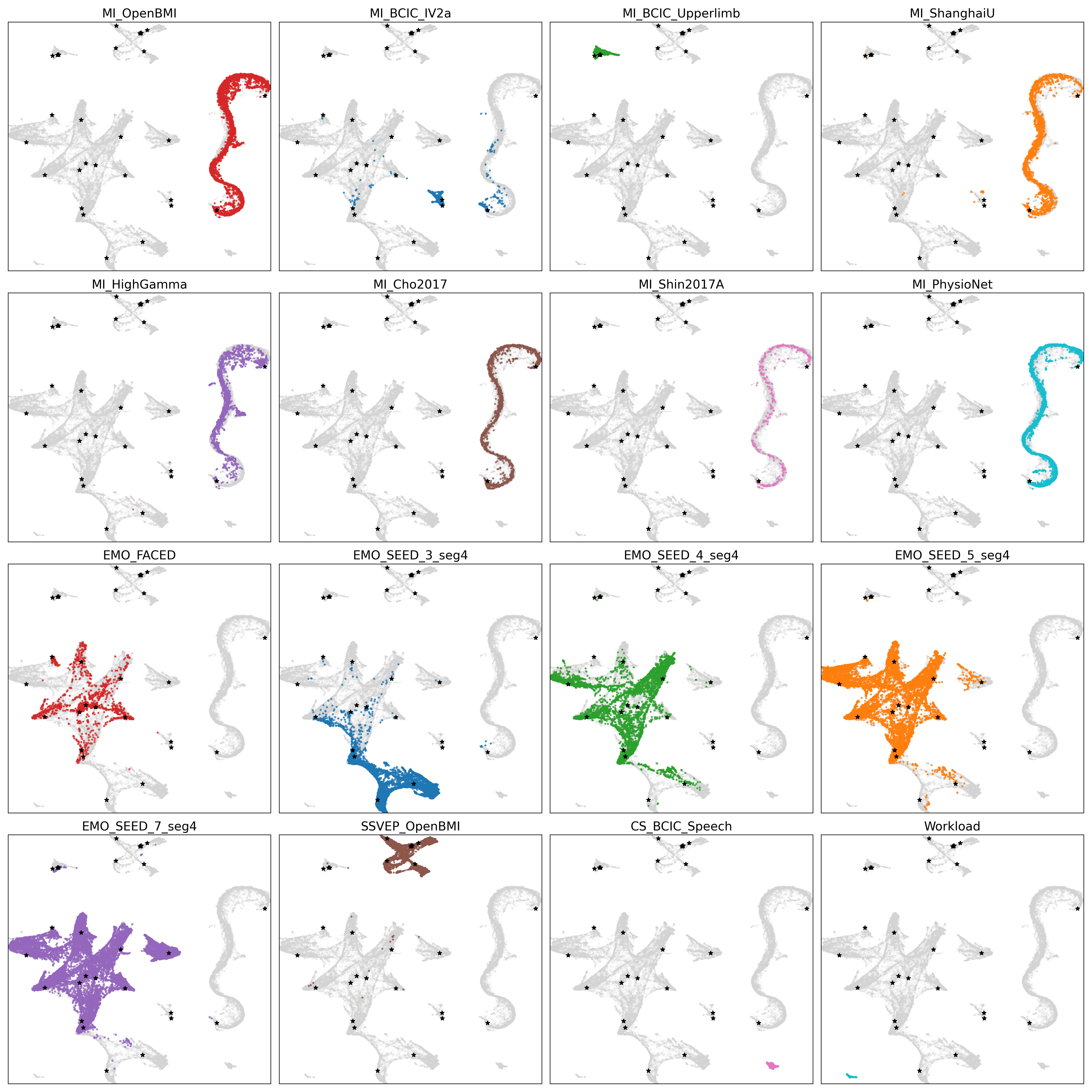}
    \caption{\textcolor{black}{Per-dataset views of the shared UMAP projection. Each panel highlights one downstream dataset in color against the embeddings of all datasets in gray; stars indicate textual class prototypes.}}
    \label{fig:umap_grid}
\end{figure*}

\textcolor{black}{Figures~\ref{fig:umap_global} and~\ref{fig:umap_grid} visualize EEG embeddings and textual class prototypes using UMAP. Motor-related datasets occupy broadly similar regions, while emotion datasets share another region of the projected space. Several textual prototypes appear close to their associated EEG embeddings, suggesting qualitative correspondence between EEG and language representations.}

\textcolor{black}{These visualizations describe representations of datasets included in instruction tuning. Their organization may reflect both EEG features and the shared instruction and label conditioning. Consequently, the projections alone do not establish generalization to unseen datasets or rule out dataset-specific memorization. Transfer to held-out datasets is evaluated separately in Appendix~\ref{appendix:heldout_zeroshot}.}

\subsection{Ablation Studies}
\label{appendix:ablations}
\textcolor{black}{For OpenBMI-MI in Tables~\ref{tab:masking_ablation}--\ref{tab:iqf_alt}, Kappa is derived from the displayed balanced accuracy using $\kappa=2\,\mathrm{B.Acc}-1$, since the test set has equal numbers of left- and right-hand trials. Where reported, its standard deviation is twice that of B.Acc.}
\subsubsection{Ablation of STR Masking Strategies}
\label{appendix:masking_ablation}
\textcolor{black}{Table~\ref{tab:masking_ablation} compares four STR masking configurations. Frequency masking alone achieves the highest B-Acc and Kappa on OpenBMI-MI, whereas random temporal masking combined with causal masking performs best on SEED. Retaining all three strategies gives the lowest values on both datasets among the configurations tested; these results do not support a consistent benefit from combining all three.}

\begin{table}[H]
\centering
\caption{Ablation of STR masking strategies. Values are B-Acc\,/\,Kappa.}
\label{tab:masking_ablation}
\renewcommand{\arraystretch}{1.1}
\resizebox{0.5\columnwidth}{!}{%
\begin{tabular}{c c c | c c}
\toprule
\textbf{Freq.} & \textbf{Rand.} & \textbf{Causal} & \textbf{OpenBMI-MI} & \textbf{SEED} \\
\midrule
\checkmark &  & & 0.8078 / \textcolor{black}{0.6156} & 0.6882 / 0.5487 \\
\checkmark & \checkmark & & 0.7896 / \textcolor{black}{0.5792} & 0.6771 / 0.5298 \\
 & \checkmark & \checkmark & 0.8033 / \textcolor{black}{0.6066} & 0.6957 / 0.5499 \\
\checkmark & \checkmark & \checkmark & 0.7851 / \textcolor{black}{0.5702} & \textcolor{black}{0.6760 / 0.5169} \\
\bottomrule
\end{tabular}}
\end{table}

\subsubsection{Analysis of Learnable Queries in IQF}
\label{appendix:query_num}
To better understand how the IQF influences our model performance, we vary the number of learnable queries $N_q \in \{4, 8, 12\}$ and evaluate model performance under the multi-task instruction tuning setting. As shown in Table~\ref{tab:query_num_ablation}, accuracy improves substantially from $N_q = 4$ to $N_q = 8$, but saturates when increasing $N_q$ to $12$. This indicates that only a small set of queries is sufficient to capture instruction-conditioned EEG patterns and that adding more queries yields diminishing returns. These results support our design that the queries function as compact latent probes rather than requiring large capacity.

\begin{table*}[h]
\centering
\caption{Direct inference ablation study on the number of learnable queries $N_q$ evaluated on two datasets using Balanced Accuracy and Cohen's Kappa.}
\label{tab:query_num_ablation}
\resizebox{0.85\textwidth}{!}{
\begin{tabular}{c|cc|cc}
\toprule
\multirow{2}{*}{\textbf{Number of Queries $N_q$}} &
\multicolumn{2}{c|}{\textbf{MI-OpenBMI}} &
\multicolumn{2}{c}{\textbf{SEED}} \\
& \textcolor{black}{\textbf{B.Acc}} & \textbf{Kappa}
& \textcolor{black}{\textbf{B.Acc}} & \textbf{Kappa} \\
\midrule
4   & 0.6913 $\pm$ 0.0025 & \textcolor{black}{0.3826 $\pm$ 0.0050} & 0.6896 $\pm$ 0.0071 & 0.5358 $\pm$ 0.0184 \\
8   & 0.7278 $\pm$ 0.0026 & \textcolor{black}{0.4556 $\pm$ 0.0052} & 0.7011 $\pm$ 0.0068 & 0.5543 $\pm$ 0.0176 \\
12  & 0.7213 $\pm$ 0.0025 & \textcolor{black}{0.4426 $\pm$ 0.0050} & 0.7004 $\pm$ 0.0069 & 0.5536 $\pm$ 0.0175 \\
\bottomrule
\end{tabular}
}
\end{table*}

\subsubsection{Ablation of IQF Cross-modal Interaction}
\label{appendix:iqf_alt}
The IQF couples instruction semantics with EEG embeddings through two design choices: (i) Feature-wise Linear Modulation (FiLM) and (ii) query-based cross-attention. To verify that this combination is non-trivial, we compare it against four alternative cross-modal interaction strategies while keeping the encoder, training schedule, and target prototypes identical:
\begin{itemize}[noitemsep,topsep=2pt,leftmargin=*]
\item \textbf{Concat}: the instruction embedding is broadcast and concatenated to each EEG token along the feature dimension; aggregation is performed by a simple MLP pooling head.
\item \textbf{FiLM-only}: instruction-conditioned FiLM is retained but learnable queries and cross-attention are removed; modulated tokens are mean-pooled before alignment.
\item \textbf{Cross-attention only}: FiLM is removed; learnable queries cross-attend to unconditioned EEG tokens.
\end{itemize}
Table~\ref{tab:iqf_alt} reports balanced accuracy and Kappa on OpenBMI-MI and SEED under multi-task instruction tuning. Each component of IQF contributes complementary gains, and the full design (FiLM + cross-attention with learnable queries) achieves the best performance, confirming that instruction conditioning and query-based bottlenecking are both required.

\begin{table}[h]
\centering
\caption{Ablation of cross-modal interaction strategies in IQF, evaluated under multi-task instruction tuning.}
\label{tab:iqf_alt}
\renewcommand{\arraystretch}{1.05}
\resizebox{0.8\textwidth}{!}{%
\begin{tabular}{l cc cc}
\toprule
\multirow{2}{*}{\textbf{Variant}} &
\multicolumn{2}{c}{\textbf{OpenBMI-MI}} &
\multicolumn{2}{c}{\textbf{SEED}} \\
\cmidrule(lr){2-3}\cmidrule(lr){4-5}
& \textcolor{black}{B.Acc} & Kappa & \textcolor{black}{B.Acc} & Kappa \\
\midrule
Concat                  & 0.6897 $\pm$ 0.0028 & \textcolor{black}{0.3794 $\pm$ 0.0056} & 0.6658 $\pm$ 0.0073 & 0.4976 $\pm$ 0.0189 \\
FiLM-only               & 0.7052 $\pm$ 0.0026 & \textcolor{black}{0.4104 $\pm$ 0.0052} & 0.6817 $\pm$ 0.0070 & 0.5239 $\pm$ 0.0182 \\
Cross-attention only    & 0.7136 $\pm$ 0.0025 & \textcolor{black}{0.4272 $\pm$ 0.0050} & 0.6919 $\pm$ 0.0069 & 0.5412 $\pm$ 0.0178 \\
\midrule
\textbf{IQF (Ours)}     & \textbf{0.7278 $\pm$ 0.0026} & \textcolor{black}{\textbf{0.4556 $\pm$ 0.0052}} & \textbf{0.7011 $\pm$ 0.0068} & \textbf{0.5543 $\pm$ 0.0176} \\
\bottomrule
\end{tabular}}
\end{table}

\subsection{Robustness and Generalization}
\label{appendix:robustness}
\subsubsection{Zero-Shot Generalization to Held-Out Datasets}
\label{appendix:heldout_zeroshot}
To test whether instruction-conditioned alignment transfers beyond the datasets used for model development, we perform direct inference on Dreyer2023 and Weibo2014, two motor-imagery datasets excluded from both pretraining and instruction tuning. No model adaptation or labeled EEG samples from either dataset are used. We compare three inference conditions: no instruction, a task-level instruction, and a task instruction augmented with the target class names. Results are reported for checkpoints trained with MPNet-base and Qwen3-Embedding-4B as the frozen text encoder.

\begin{figure*}[t]
    \centering
    \includegraphics[width=0.95\textwidth]{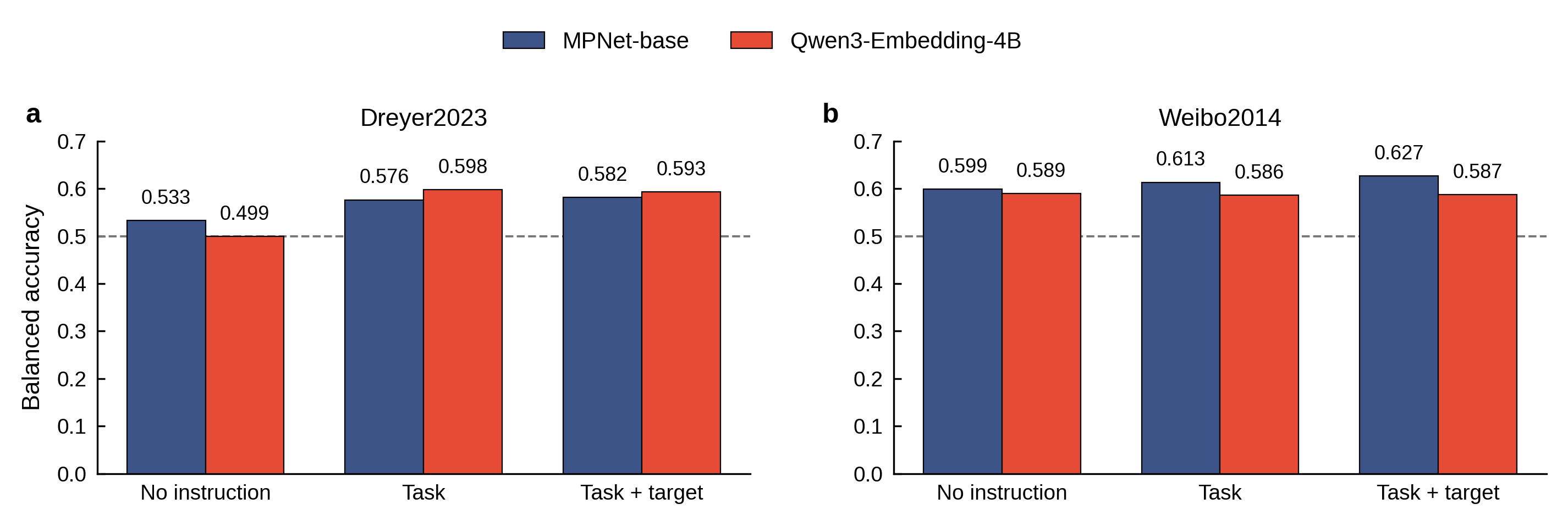}
    \caption{Zero-shot direct inference on held-out motor-imagery datasets. Balanced accuracy on Dreyer2023 and Weibo2014 using MPNet-base or Qwen3-Embedding-4B as the frozen text encoder. The models are evaluated without adaptation under three levels of semantic guidance: no instruction, task instruction, and task instruction with target class names. The dashed horizontal line denotes chance performance (0.5).}
    \label{fig:zero_shot_heldout}
\end{figure*}

As shown in Fig.~\ref{fig:zero_shot_heldout}, instructions substantially improve performance on Dreyer2023: balanced accuracy increases from 0.533 to 0.582 with MPNet-base and from 0.499 to 0.593 with Qwen3-Embedding-4B when task and target information are supplied. On Weibo2014, MPNet-base improves from 0.599 to 0.627, whereas Qwen3-Embedding-4B remains approximately unchanged (0.589 versus 0.587). Averaged across the two held-out datasets, the task-and-target condition improves balanced accuracy from 0.566 to 0.604 for MPNet-base and from 0.544 to 0.590 for Qwen3-Embedding-4B. These results show that language guidance can support zero-shot transfer, particularly when the unconditioned representation is weak, while its benefit remains dependent on the dataset and text-embedding geometry.

\begin{figure*}[!t]
    \centering
    \includegraphics[width=\textwidth]{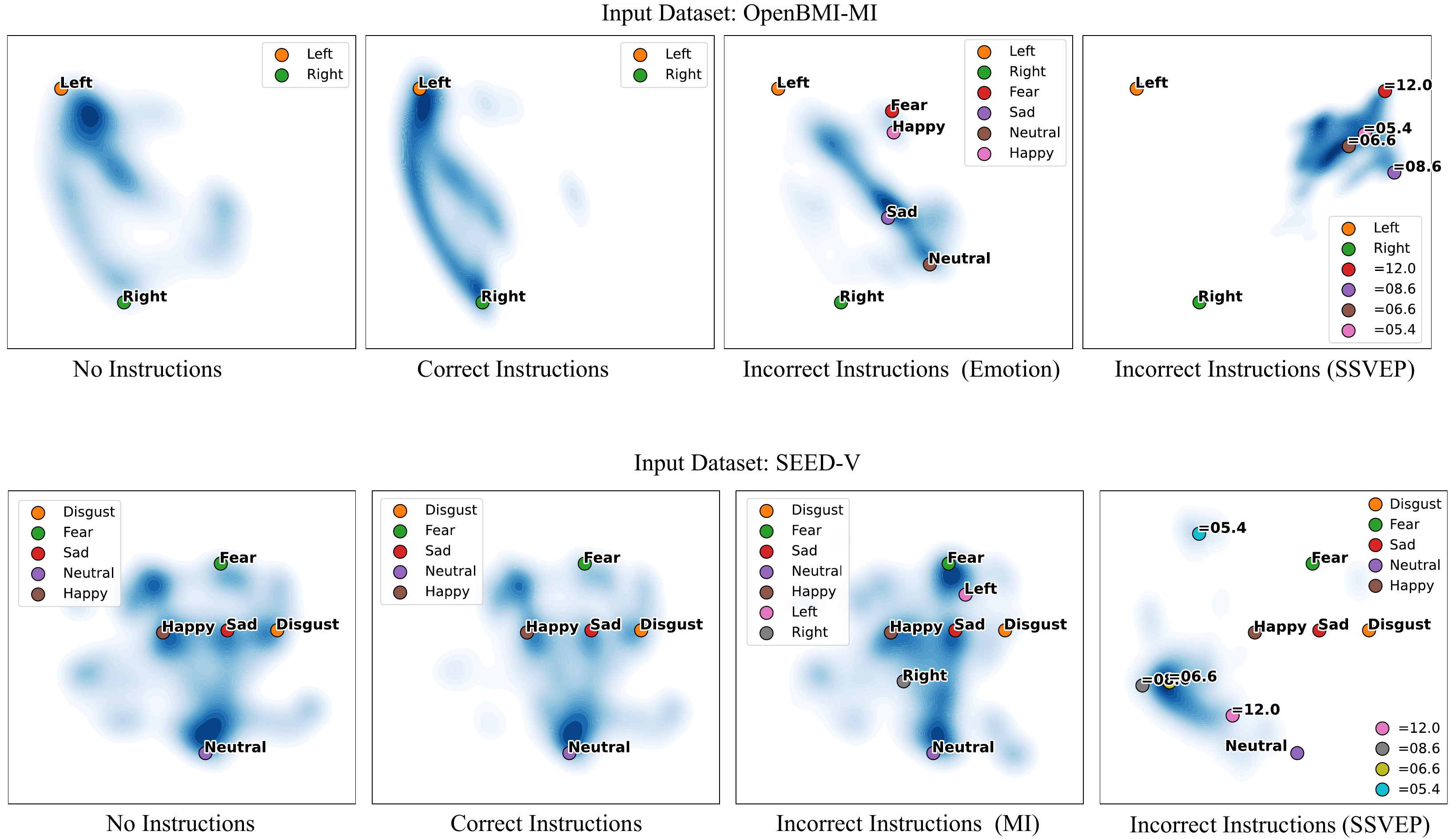}
    \caption{Comparison of KDE visualization of features between incorrect and correct instructions.}
    \label{fig:mislead}
\end{figure*}

\subsubsection{Effect of Incorrect Instructions}
\label{appendix:incorrect_instructions}
To further examine the role of language guidance, we analyze cases where the model is deliberately given misleading instructions that do not match the underlying EEG dataset. Fig.~\ref{fig:mislead} shows examples on OpenBMI-MI and SEED-V datasets.

When provided with correct instructions, the learned feature spaces become more structured, with compact intra-class clusters and clearer inter-class separation. However, when misleading instructions are introduced, the feature space is distorted toward the semantics of the given instruction rather than the ground-truth task. For example, MI data conditioned on emotion-related instructions form clusters resembling affective categories, and SEED-V data prompted with MI or SSVEP instructions are reorganized into motor or frequency-based groupings.
These results emphasize the strong controllability of our model through natural language. While correct instructions enhance discriminability, misleading instructions actively reshape the representation space according to the semantic prior they provide. This highlights both the power and sensitivity of instruction-conditioned alignment in EEG-FMs.
\begin{table}[!t]
\centering
\caption{Direct inference evaluation across instruction sentences for motor imagery and emotion tasks. \textcolor{black}{Summary rows report the mean and sample standard deviation across the ten listed paraphrases.}}
\label{tab:paraphrase}
\renewcommand{\arraystretch}{1.0}
\scriptsize
\begin{subtable}{1.0\textwidth}
\centering
\caption{Motor Imagery Task}
\begin{adjustbox}{width=\textwidth}
\begin{tabular}{p{0.40\textwidth}|cc|cc}
\toprule
\multirow{2}{*}{\textbf{Instruction Sentence}}
& \multicolumn{2}{c|}{\textbf{MI-OpenBMI}}
& \multicolumn{2}{c}{\textbf{MI-Cho2017}} \\
\cmidrule(lr){2-3}\cmidrule(lr){4-5}
& B.Acc & Kappa & B.Acc & Kappa \\
\midrule
This is a motor imagery decoding task                   & 0.7190 & 0.4379 & 0.7356 & 0.4711 \\
Decode motor imagery                                    & 0.7206 & 0.4413 & 0.7333 & 0.4667 \\
Please decode the motor imagery                         & 0.7208 & 0.4417 & 0.7361 & 0.4722 \\
Classify the motor imagery task                         & 0.7225 & 0.4450 & 0.7339 & 0.4678 \\
Identify the motor imagery class                        & 0.7223 & 0.4446 & 0.7356 & 0.4711 \\
Predict the type of imagined movement                   & 0.7206 & 0.4413 & 0.7328 & 0.4656 \\
Determine which motor imagery action is performed       & 0.7244 & 0.4487 & 0.7344 & 0.4689 \\
Infer the imagined motor action from the EEG            & 0.7250 & 0.4500 & 0.7306 & 0.4611 \\
Which motor imagery category does this trial belong to? & 0.7165 & 0.4329 & 0.7344 & 0.4689 \\
Analyze this EEG and classify the motor imagery         & 0.7208 & 0.4417 & 0.7306 & 0.4611 \\
\midrule
\textbf{Average $\pm$ STD} & 0.7213 $\pm$ 0.0025 & 0.4425 $\pm$ 0.0050 & 0.7337 $\pm$ 0.0019 & 0.4675 $\pm$ 0.0039 \\
\bottomrule
\end{tabular}
\end{adjustbox}
\end{subtable}

\vspace{0.3cm}
\begin{subtable}{1.0\textwidth}
\centering
\caption{Emotion Recognition Task}
\begin{adjustbox}{width=\textwidth}
\begin{tabular}{p{0.40\textwidth}|cc|cc}
\toprule
\multirow{2}{*}{\textbf{Instruction Sentence}}
& \multicolumn{2}{c|}{\textbf{SEED}}
& \multicolumn{2}{c}{\textbf{SEED-IV}} \\
\cmidrule(lr){2-3}\cmidrule(lr){4-5}
& B.Acc & Kappa & B.Acc & Kappa \\
\midrule
This is an emotion recognition task             & 0.5602 & 0.3481 & 0.3779 & 0.1931 \\
Decode emotional states                         & 0.5663 & 0.3581 & 0.3835 & 0.2039 \\
Please decode the emotional states              & 0.5643 & 0.3552 & 0.3838 & 0.2041 \\
Identify the emotional state                    & 0.5659 & 0.3574 & 0.3838 & 0.2048 \\
Classify the emotion represented in the signal  & 0.5663 & 0.3579 & 0.3811 & 0.2007 \\
Predict the emotion category                    & 0.5656 & 0.3568 & 0.3831 & 0.2031 \\
Determine the emotional condition from this EEG & 0.5646 & 0.3553 & 0.3818 & 0.2008 \\
Infer the participant emotional state           & 0.5667 & 0.3588 & 0.3836 & 0.2030 \\
What emotion does this EEG signal reflect?      & 0.5655 & 0.3567 & 0.3818 & 0.2028 \\
Analyze the EEG and classify the emotion        & 0.5662 & 0.3580 & 0.3818 & 0.2025 \\
\midrule
\textbf{Average $\pm$ STD} & \textcolor{black}{0.5652 $\pm$ 0.0019} & \textcolor{black}{0.3562 $\pm$ 0.0031} & 0.3822 $\pm$ 0.0018 & 0.2019 $\pm$ 0.0034 \\
\bottomrule
\end{tabular}
\end{adjustbox}
\end{subtable}
\end{table}

\subsubsection{Instruction Paraphrase Robustness}
\label{appendix:paraphrase}
To evaluate whether LEAF depends on the exact linguistic phrasing of the task instruction, we conducted a paraphrase robustness study. For each task, we designed 10 semantically equivalent instructions that vary in grammatical structure (declarative, imperative, interrogative), lexical choice (e.g., ``decode'', ``classify'', ``identify'', ``infer''), and formality, while preserving the intended task meaning. No hyperparameters or training settings were changed; only the instruction string passed to the frozen text encoder was varied. As shown in Table~\ref{tab:paraphrase}, B.Acc and Kappa remain tightly concentrated across all paraphrases on both motor imagery and emotion datasets (e.g., MI-OpenBMI B.Acc STD\,=\,0.0025; SEED B.Acc STD\,=\,\textcolor{black}{0.0019}), demonstrating that LEAF relies on the underlying semantic content of the instruction rather than the precise linguistic template.

\section{Discussion}
\label{appendix:discussion}
\subsection{Limitations and Future Work}
\label{appendix:limitations}
While LEAF aligns EEG embeddings with textual label prototypes in a shared EEG--language space, our evaluation has so far focused on label sets encountered during instruction tuning, and we have not yet systematically probed inference with entirely novel class names. \textcolor{black}{The direct inference regime uses textual prototypes for decoding and is therefore in principle compatible with open-vocabulary decoding, where unseen class names can serve as prototypes at inference time. The task-specific finetuning regime uses a learned classification head with a fixed output label set.} A thorough empirical study of this capability, including the conditions under which it succeeds or degrades, remains a promising direction for future work.
\subsection{Broader Impacts}
\label{appendix:broader_impacts}
LEAF advances non-invasive BCI technology by enabling generalizable EEG decoding across diverse tasks, with direct benefits for individuals with motor disabilities, speech impairments, and neurological conditions. By unifying EEG representations through natural language, LEAF lowers the barrier to developing accessible assistive systems. While these applications are broadly beneficial, careful attention must be paid to mitigating risks such as algorithmic bias across subject demographics and patient data privacy in clinical usage.

\clearpage
\section*{NeurIPS Paper Checklist}

\begin{enumerate}

\item {\bf Claims}
    \item[] Question: Do the main claims made in the abstract and introduction accurately reflect the paper's contributions and scope?
    \item[] Answer: \answerYes{}
    \item[] Justification: The abstract and introduction clearly state all four contributions (STR pretraining, IQF alignment module, state-of-the-art results on 12 of 16 datasets, and the first empirical demonstration that task instructions serve as semantic priors for EEG representations), all of which are fully supported by the experimental results in Section 3.
    \item[] Guidelines:
    \begin{itemize}
        \item The answer \answerNA{} means that the abstract and introduction do not include the claims made in the paper.
        \item The abstract and/or introduction should clearly state the claims made, including the contributions made in the paper and important assumptions and limitations. A \answerNo{} or \answerNA{} answer to this question will not be perceived well by the reviewers. 
        \item The claims made should match theoretical and experimental results, and reflect how much the results can be expected to generalize to other settings. 
        \item It is fine to include aspirational goals as motivation as long as it is clear that these goals are not attained by the paper. 
    \end{itemize}

\item {\bf Limitations}
    \item[] Question: Does the paper discuss the limitations of the work performed by the authors?
    \item[] Answer: \answerYes{}
    \item[] Justification: A Limitations and Future Work section is included in the Appendix, noting that LEAF has not yet been systematically evaluated on entirely unseen class names and identifying open-vocabulary decoding as a promising direction for future work.
    \item[] Guidelines:
    \begin{itemize}
        \item The answer \answerNA{} means that the paper has no limitation while the answer \answerNo{} means that the paper has limitations, but those are not discussed in the paper. 
        \item The authors are encouraged to create a separate ``Limitations'' section in their paper.
        \item The paper should point out any strong assumptions and how robust the results are to violations of these assumptions (e.g., independence assumptions, noiseless settings, model well-specification, asymptotic approximations only holding locally). The authors should reflect on how these assumptions might be violated in practice and what the implications would be.
        \item The authors should reflect on the scope of the claims made, e.g., if the approach was only tested on a few datasets or with a few runs. In general, empirical results often depend on implicit assumptions, which should be articulated.
        \item The authors should reflect on the factors that influence the performance of the approach. For example, a facial recognition algorithm may perform poorly when image resolution is low or images are taken in low lighting. Or a speech-to-text system might not be used reliably to provide closed captions for online lectures because it fails to handle technical jargon.
        \item The authors should discuss the computational efficiency of the proposed algorithms and how they scale with dataset size.
        \item If applicable, the authors should discuss possible limitations of their approach to address problems of privacy and fairness.
        \item While the authors might fear that complete honesty about limitations might be used by reviewers as grounds for rejection, a worse outcome might be that reviewers discover limitations that aren't acknowledged in the paper. The authors should use their best judgment and recognize that individual actions in favor of transparency play an important role in developing norms that preserve the integrity of the community. Reviewers will be specifically instructed to not penalize honesty concerning limitations.
    \end{itemize}

\item {\bf Theory assumptions and proofs}
    \item[] Question: For each theoretical result, does the paper provide the full set of assumptions and a complete (and correct) proof?
    \item[] Answer: \answerNA{}
    \item[] Justification: LEAF is an empirical EEG foundation model; the paper focuses on architecture design, pretraining objectives, and downstream evaluation rather than theoretical analysis. No formal theorems or proofs are claimed.
    \item[] Guidelines:
    \begin{itemize}
        \item The answer \answerNA{} means that the paper does not include theoretical results. 
        \item All the theorems, formulas, and proofs in the paper should be numbered and cross-referenced.
        \item All assumptions should be clearly stated or referenced in the statement of any theorems.
        \item The proofs can either appear in the main paper or the supplemental material, but if they appear in the supplemental material, the authors are encouraged to provide a short proof sketch to provide intuition. 
        \item Inversely, any informal proof provided in the core of the paper should be complemented by formal proofs provided in appendix or supplemental material.
        \item Theorems and Lemmas that the proof relies upon should be properly referenced. 
    \end{itemize}

    \item {\bf Experimental result reproducibility}
    \item[] Question: Does the paper fully disclose all the information needed to reproduce the main experimental results of the paper to the extent that it affects the main claims and/or conclusions of the paper (regardless of whether the code and data are provided or not)?
    \item[] Answer: \answerYes{}
    \item[] Justification: Full architecture specifications, hyperparameters, training configuration, dataset preprocessing, and train/validation/test split strategies are provided in Appendices~\ref{appendix:experimental_details} and~\ref{appendix:implementation}. Code and pre-trained weights will be released to further support reproducibility.
    \item[] Guidelines:
    \begin{itemize}
        \item The answer \answerNA{} means that the paper does not include experiments.
        \item If the paper includes experiments, a \answerNo{} answer to this question will not be perceived well by the reviewers: Making the paper reproducible is important, regardless of whether the code and data are provided or not.
        \item If the contribution is a dataset and\slash or model, the authors should describe the steps taken to make their results reproducible or verifiable. 
        \item Depending on the contribution, reproducibility can be accomplished in various ways. For example, if the contribution is a novel architecture, describing the architecture fully might suffice, or if the contribution is a specific model and empirical evaluation, it may be necessary to either make it possible for others to replicate the model with the same dataset, or provide access to the model. In general. releasing code and data is often one good way to accomplish this, but reproducibility can also be provided via detailed instructions for how to replicate the results, access to a hosted model (e.g., in the case of a large language model), releasing of a model checkpoint, or other means that are appropriate to the research performed.
        \item While NeurIPS does not require releasing code, the conference does require all submissions to provide some reasonable avenue for reproducibility, which may depend on the nature of the contribution. For example
        \begin{enumerate}
            \item If the contribution is primarily a new algorithm, the paper should make it clear how to reproduce that algorithm.
            \item If the contribution is primarily a new model architecture, the paper should describe the architecture clearly and fully.
            \item If the contribution is a new model (e.g., a large language model), then there should either be a way to access this model for reproducing the results or a way to reproduce the model (e.g., with an open-source dataset or instructions for how to construct the dataset).
            \item We recognize that reproducibility may be tricky in some cases, in which case authors are welcome to describe the particular way they provide for reproducibility. In the case of closed-source models, it may be that access to the model is limited in some way (e.g., to registered users), but it should be possible for other researchers to have some path to reproducing or verifying the results.
        \end{enumerate}
    \end{itemize}

\item {\bf Open access to data and code}
    \item[] Question: Does the paper provide open access to the data and code, with sufficient instructions to faithfully reproduce the main experimental results, as described in supplemental material?
    \item[] Answer: \answerYes{}
    \item[] Justification: All EEG datasets used in this paper are publicly available and properly cited. The authors provide full access to the code at \url{https://github.com/Jiang-Muyun/LEAF}.
    \item[] Guidelines:
    \begin{itemize}
        \item The answer \answerNA{} means that paper does not include experiments requiring code.
        \item Please see the NeurIPS code and data submission guidelines (\url{https://neurips.cc/public/guides/CodeSubmissionPolicy}) for more details.
        \item While we encourage the release of code and data, we understand that this might not be possible, so \answerNo{} is an acceptable answer. Papers cannot be rejected simply for not including code, unless this is central to the contribution (e.g., for a new open-source benchmark).
        \item The instructions should contain the exact command and environment needed to run to reproduce the results. See the NeurIPS code and data submission guidelines (\url{https://neurips.cc/public/guides/CodeSubmissionPolicy}) for more details.
        \item The authors should provide instructions on data access and preparation, including how to access the raw data, preprocessed data, intermediate data, and generated data, etc.
        \item The authors should provide scripts to reproduce all experimental results for the new proposed method and baselines. If only a subset of experiments are reproducible, they should state which ones are omitted from the script and why.
        \item At submission time, to preserve anonymity, the authors should release anonymized versions (if applicable).
        \item Providing as much information as possible in supplemental material (appended to the paper) is recommended, but including URLs to data and code is permitted.
    \end{itemize}

\item {\bf Experimental setting/details}
    \item[] Question: Does the paper specify all the training and test details (e.g., data splits, hyperparameters, how they were chosen, type of optimizer) necessary to understand the results?
    \item[] Answer: \answerYes{}
    \item[] Justification: The paper fully specifies the experimental setup, including dataset splits, hyperparameters (optimizer, learning rate schedule, batch size), and preprocessing details. Key settings are summarized in Section 3.2, with comprehensive specifications provided in Appendices~\ref{appendix:experimental_details} and~\ref{appendix:implementation}.
    \item[] Guidelines:
    \begin{itemize}
        \item The answer \answerNA{} means that the paper does not include experiments.
        \item The experimental setting should be presented in the core of the paper to a level of detail that is necessary to appreciate the results and make sense of them.
        \item The full details can be provided either with the code, in appendix, or as supplemental material.
    \end{itemize}

\item {\bf Experiment statistical significance}
    \item[] Question: Does the paper report error bars suitably and correctly defined or other appropriate information about the statistical significance of the experiments?
    \item[] Answer: \answerYes{}
    \item[] Justification: Balanced accuracy and Cohen's Kappa are adopted as primary metrics across all 16 datasets. All results in Table 1 are reported as mean $\pm$ standard deviation over three independent random seeds, consistent with the evaluation protocol of all baselines, capturing variability due to random initialization.
    \item[] Guidelines:
    \begin{itemize}
        \item The answer \answerNA{} means that the paper does not include experiments.
        \item The authors should answer \answerYes{} if the results are accompanied by error bars, confidence intervals, or statistical significance tests, at least for the experiments that support the main claims of the paper.
        \item The factors of variability that the error bars are capturing should be clearly stated (for example, train/test split, initialization, random drawing of some parameter, or overall run with given experimental conditions).
        \item The method for calculating the error bars should be explained (closed form formula, call to a library function, bootstrap, etc.)
        \item The assumptions made should be given (e.g., Normally distributed errors).
        \item It should be clear whether the error bar is the standard deviation or the standard error of the mean.
        \item It is OK to report 1-sigma error bars, but one should state it. The authors should preferably report a 2-sigma error bar than state that they have a 96\% CI, if the hypothesis of Normality of errors is not verified.
        \item For asymmetric distributions, the authors should be careful not to show in tables or figures symmetric error bars that would yield results that are out of range (e.g., negative error rates).
        \item If error bars are reported in tables or plots, the authors should explain in the text how they were calculated and reference the corresponding figures or tables in the text.
    \end{itemize}

\item {\bf Experiments compute resources}
    \item[] Question: For each experiment, does the paper provide sufficient information on the computer resources (type of compute workers, memory, time of execution) needed to reproduce the experiments?
    \item[] Answer: \answerYes{}
    \item[] Justification: Compute resources are detailed in Appendix~\ref{appendix:compute}, including GPU model (2$\times$NVIDIA RTX PRO 5000, 48\,GB each), CPU, memory, and OS specifications. Training times are also reported: approximately 4 hours for pre-training and 2 hours for instruction tuning.
    \item[] Guidelines:
    \begin{itemize}
        \item The answer \answerNA{} means that the paper does not include experiments.
        \item The paper should indicate the type of compute workers CPU or GPU, internal cluster, or cloud provider, including relevant memory and storage.
        \item The paper should provide the amount of compute required for each of the individual experimental runs as well as estimate the total compute. 
        \item The paper should disclose whether the full research project required more compute than the experiments reported in the paper (e.g., preliminary or failed experiments that didn't make it into the paper). 
    \end{itemize}
    
\item {\bf Code of ethics}
    \item[] Question: Does the research conducted in the paper conform, in every respect, with the NeurIPS Code of Ethics \url{https://neurips.cc/public/EthicsGuidelines}?
    \item[] Answer: \answerYes{}
    \item[] Justification: This research uses only publicly available EEG datasets and introduces no new data collection involving human subjects. The work conforms fully with the NeurIPS Code of Ethics.
    \item[] Guidelines:
    \begin{itemize}
        \item The answer \answerNA{} means that the authors have not reviewed the NeurIPS Code of Ethics.
        \item If the authors answer \answerNo, they should explain the special circumstances that require a deviation from the Code of Ethics.
        \item The authors should make sure to preserve anonymity (e.g., if there is a special consideration due to laws or regulations in their jurisdiction).
    \end{itemize}

\item {\bf Broader impacts}
    \item[] Question: Does the paper discuss both potential positive societal impacts and negative societal impacts of the work performed?
    \item[] Answer: \answerYes{}
    \item[] Justification: A Broader Impacts section is included in the Appendix, discussing both positive impacts (advancing assistive BCI for individuals with motor and speech disabilities) and negative risks (algorithmic bias and patient data privacy in clinical usage).
    \item[] Guidelines:
    \begin{itemize}
        \item The answer \answerNA{} means that there is no societal impact of the work performed.
        \item If the authors answer \answerNA{} or \answerNo, they should explain why their work has no societal impact or why the paper does not address societal impact.
        \item Examples of negative societal impacts include potential malicious or unintended uses (e.g., disinformation, generating fake profiles, surveillance), fairness considerations (e.g., deployment of technologies that could make decisions that unfairly impact specific groups), privacy considerations, and security considerations.
        \item The conference expects that many papers will be foundational research and not tied to particular applications, let alone deployments. However, if there is a direct path to any negative applications, the authors should point it out. For example, it is legitimate to point out that an improvement in the quality of generative models could be used to generate Deepfakes for disinformation. On the other hand, it is not needed to point out that a generic algorithm for optimizing neural networks could enable people to train models that generate Deepfakes faster.
        \item The authors should consider possible harms that could arise when the technology is being used as intended and functioning correctly, harms that could arise when the technology is being used as intended but gives incorrect results, and harms following from (intentional or unintentional) misuse of the technology.
        \item If there are negative societal impacts, the authors could also discuss possible mitigation strategies (e.g., gated release of models, providing defenses in addition to attacks, mechanisms for monitoring misuse, mechanisms to monitor how a system learns from feedback over time, improving the efficiency and accessibility of ML).
    \end{itemize}
    
\item {\bf Safeguards}
    \item[] Question: Does the paper describe safeguards that have been put in place for responsible release of data or models that have a high risk for misuse (e.g., pre-trained language models, image generators, or scraped datasets)?
    \item[] Answer: \answerNA{}
    \item[] Justification: The paper does not introduce any such risk.
    \item[] Guidelines:
    \begin{itemize}
        \item The answer \answerNA{} means that the paper poses no such risks.
        \item Released models that have a high risk for misuse or dual-use should be released with necessary safeguards to allow for controlled use of the model, for example by requiring that users adhere to usage guidelines or restrictions to access the model or implementing safety filters. 
        \item Datasets that have been scraped from the Internet could pose safety risks. The authors should describe how they avoided releasing unsafe images.
        \item We recognize that providing effective safeguards is challenging, and many papers do not require this, but we encourage authors to take this into account and make a best faith effort.
    \end{itemize}

\item {\bf Licenses for existing assets}
    \item[] Question: Are the creators or original owners of assets (e.g., code, data, models), used in the paper, properly credited and are the license and terms of use explicitly mentioned and properly respected?
    \item[] Answer: \answerYes{}
    \item[] Justification: All datasets used in this paper are publicly available academic resources and are properly cited. The baseline model implementations are used in accordance with their official releases.
    \item[] Guidelines:
    \begin{itemize}
        \item The answer \answerNA{} means that the paper does not use existing assets.
        \item The authors should cite the original paper that produced the code package or dataset.
        \item The authors should state which version of the asset is used and, if possible, include a URL.
        \item The name of the license (e.g., CC-BY 4.0) should be included for each asset.
        \item For scraped data from a particular source (e.g., website), the copyright and terms of service of that source should be provided.
        \item If assets are released, the license, copyright information, and terms of use in the package should be provided. For popular datasets, \url{paperswithcode.com/datasets} has curated licenses for some datasets. Their licensing guide can help determine the license of a dataset.
        \item For existing datasets that are re-packaged, both the original license and the license of the derived asset (if it has changed) should be provided.
        \item If this information is not available online, the authors are encouraged to reach out to the asset's creators.
    \end{itemize}

\item {\bf New assets}
    \item[] Question: Are new assets introduced in the paper well documented and is the documentation provided alongside the assets?
    \item[] Answer: \answerYes{}
    \item[] Justification: All EEG datasets used in this paper are publicly available and properly cited. The authors provide full access to the code at \url{https://github.com/Jiang-Muyun/LEAF}.
    \item[] Guidelines:
    \begin{itemize}
        \item The answer \answerNA{} means that the paper does not release new assets.
        \item Researchers should communicate the details of the dataset\slash code\slash model as part of their submissions via structured templates. This includes details about training, license, limitations, etc. 
        \item The paper should discuss whether and how consent was obtained from people whose asset is used.
        \item At submission time, remember to anonymize your assets (if applicable). You can either create an anonymized URL or include an anonymized zip file.
    \end{itemize}

\item {\bf Crowdsourcing and research with human subjects}
    \item[] Question: For crowdsourcing experiments and research with human subjects, does the paper include the full text of instructions given to participants and screenshots, if applicable, as well as details about compensation (if any)?
    \item[] Answer: \answerNA{}
    \item[] Justification: This paper does not involve any crowdsourcing or new human subject data collection. All EEG data used are from existing publicly available datasets collected by the original dataset authors.
    \item[] Guidelines:
    \begin{itemize}
        \item The answer \answerNA{} means that the paper does not involve crowdsourcing nor research with human subjects.
        \item Including this information in the supplemental material is fine, but if the main contribution of the paper involves human subjects, then as much detail as possible should be included in the main paper. 
        \item According to the NeurIPS Code of Ethics, workers involved in data collection, curation, or other labor should be paid at least the minimum wage in the country of the data collector. 
    \end{itemize}

\item {\bf Institutional review board (IRB) approvals or equivalent for research with human subjects}
    \item[] Question: Does the paper describe potential risks incurred by study participants, whether such risks were disclosed to the subjects, and whether Institutional Review Board (IRB) approvals (or an equivalent approval/review based on the requirements of your country or institution) were obtained?
    \item[] Answer: \answerNA{}
    \item[] Justification: No new human subject data was collected in this work. All datasets used were gathered and ethically approved by their original authors prior to public release.
    \item[] Guidelines:
    \begin{itemize}
        \item The answer \answerNA{} means that the paper does not involve crowdsourcing nor research with human subjects.
        \item Depending on the country in which research is conducted, IRB approval (or equivalent) may be required for any human subjects research. If you obtained IRB approval, you should clearly state this in the paper. 
        \item We recognize that the procedures for this may vary significantly between institutions and locations, and we expect authors to adhere to the NeurIPS Code of Ethics and the guidelines for their institution. 
        \item For initial submissions, do not include any information that would break anonymity (if applicable), such as the institution conducting the review.
    \end{itemize}

\item {\bf Declaration of LLM usage}
    \item[] Question: Does the paper describe the usage of LLMs if it is an important, original, or non-standard component of the core methods in this research? Note that if the LLM is used only for writing, editing, or formatting purposes and does \emph{not} impact the core methodology, scientific rigor, or originality of the research, declaration is not required.
    \item[] Answer: \answerYes{}
    \item[] Justification: The paper describes the use of frozen pretrained text encoders for instruction and class-label embeddings. The main model uses all-mpnet-base-v2; Qwen3-Embedding-4B is additionally evaluated in text-encoder comparisons and zero-shot transfer experiments on held-out datasets. These encoders provide fixed text representations and are not fine-tuned. An LLM was also used for writing assistance.
    \item[] Guidelines:
    \begin{itemize}
        \item The answer \answerNA{} means that the core method development in this research does not involve LLMs as any important, original, or non-standard components.
        \item Please refer to our LLM policy in the NeurIPS handbook for what should or should not be described.
    \end{itemize}

\end{enumerate}

\end{document}